\relax
\documentclass[letterpaper]{article} 
\usepackage[final]{neurips_2021}
\usepackage{times}  
\usepackage{helvet}  
\usepackage{courier}  
\usepackage[hyphens]{url}  
\usepackage{graphicx} 
\usepackage{natbib}  
\usepackage{caption} 
\usepackage{multirow}
\usepackage{graphicx}
\usepackage{booktabs}
\usepackage{amsmath}

\usepackage{xspace}   

\newcommand{\yaxis}{$y$-axis\xspace}

\usepackage{color}
\definecolor{orange}{RGB}{255,127,0}
\definecolor{brown}{RGB}{150,70,0}
\definecolor{green}{RGB}{127,255,127}
\definecolor{darkgreen}{RGB}{0,127,0}
\definecolor{blue}{RGB}{127,127,255}
\definecolor{lightblue}{RGB}{150,150,255}
\definecolor{darkblue}{RGB}{0,0,127}
\definecolor{red}{RGB}{255,90,90}
\definecolor{violet}{RGB}{200,110,170}
\definecolor{grey}{RGB}{127,127,127}
\definecolor{pink}{RGB}{255,180,180}

\DeclareMathAlphabet{\mathbbmsl}{U}{bbm}{m}{sl}

\title{Compute and Energy Consumption Trends in \\ Deep Learning Inference}

\author{
  Radosvet Desislavov \\
  VRAIN. Universitat Politècnica de València, Spain\\
  \texttt{radegeo@inf.upv.es} \\
  \And
  Fernando Martínez-Plumed \\
  European Commission, Joint Research Centre \\
  \texttt{fernando.martinez-plumed@ec.europa.eu} \\
  VRAIN. Universitat Politècnica de València, Spain\\
  \texttt{fmartinez@dsic.upv.es} \\
  \AND
  José Hernández-Orallo \\
  VRAIN. Universitat Politècnica de València, Spain\\
  \texttt{jorallo@upv.es} \\
}

\usepackage{bibentry}

\begin{document}

\maketitle

\begin{abstract}
The progress of some AI paradigms such as deep learning is said to be linked to an exponential growth in the number of parameters. 
There are many studies corroborating these trends, but does this translate into an exponential increase in energy consumption? In order to answer this question we 
focus on inference costs rather than training costs, as the former account for most of the computing effort, solely because of the multiplicative factors. 
Also, apart from algorithmic innovations, we 
account for more specific and powerful hardware (leading to higher FLOPS) that is usually accompanied with important energy efficiency optimisations. We  also 
move the focus from the first implementation of a breakthrough paper towards the consolidated version of the techniques one or two year later. Under this distinctive and comprehensive perspective, we study relevant models in the areas of computer vision and natural language processing: for a sustained increase in performance we see a much softer growth in energy consumption than previously anticipated. The only caveat is, yet again, the multiplicative factor, as future AI increases penetration and becomes more pervasive. 
\end{abstract}

\section{Introduction}

As Deep Neural Networks (DNNs) become more widespread in all kinds of devices and situations, what is the associated energy cost? In this work we explore the evolution of different metrics of deep learning models, paying particular attention to {\em inference} computational cost and its associated energy consumption. The full impact, and its final carbon footprint, not only depends on the internalities (hardware and software directly involved in their operation) but also on the externalities (all social and economic activities around it). From the AI research community, we have more to say and do about the former. Accordingly, more effort is needed, within AI, to better account for the internalities, as we do in this paper.

In our study, we 
differentiate between training and inference. At first look it seems that training cost is higher. However, for 
deployed systems, inference costs exceed training costs,  because of the multiplicative factor of using the system many times \citep{martinez2018accounting}. Training, even if it involves repetitions, is done once but inference is done repeatedly. It is estimated that inference accounts for up to 90\% of the costs \citep{aws}. There are several studies about training computation and its environmental impact \citep{ai_and_compute,gholami2020ai_and_memory_wall,canziani2017analysis,7723730,anthony2020carbontracker,thompson2020computational} 
but there are very few focused on inference costs and their associated energy consumption.

DNNs are deployed almost everywhere \citep{balas2019handbook},  
from smartphones to automobiles, all having their own compute, temperature and battery limitations. Precisely because of this, there has been a pressure to build DNNs that are less resource demanding, even if larger DNNs usually outperform smaller ones. Alternatively to this in-device use, many larger DNNs are run on data centres, with people accessing them repeated in a transparent way, e.g., when using social networks 
\citep{park2018deep}. 
Millions of requests imply millions of inferences over the same DNN.

Many studies report that the size of neural networks is growing exponentially \citep{xu2018scaling,bianco2018benchmark}. 
However, this does not necessarily imply that the cost is also growing exponentially, as more weights could be implemented with the same amount of energy, 
mostly due to hardware specialisation but especially as the energy consumption per unit of compute is decreasing. Also, there is the question of whether the changing costs of energy and their carbon footprint \citep{europa-footprint} should be added to the equation. Finally, many studies focus on the state-of-the-art (SOTA) or the cutting-edge methods according to a given metric of performance, but many algorithmic improvements usually come in the months or few years after a new technique is introduced, in the form of {\em general use} implementations having similar results with much lower compute requirements. All these elements have been studied separately, but a more comprehensive and integrated analysis 
is necessary to properly evaluate whether the impact of AI on energy consumption and its carbon footprint is alarming or simply worrying, in order to calibrate the measures to be taken in the following years and estimate the effect in the future.

For conducting our analysis we chose two representative domains: Computer Vision (CV) and Natural Language Processing (NLP). For CV we analysed image classification models, 
and ImageNet  \citep{russakovsky2015ImageNet} more specifically, because 
there is a great quantity of historical data in this area and 
many advances in this domain are normally brought to other computer vision tasks, such as object detection, semantic segmentation, action recognition, or video classification, among others. For NLP 
we analysed results for the General Language Understanding Evaluation (GLUE) benchmark \citep{wang2019glue}, since language understanding is a core task in NLP. 

We focus our analysis on inference FLOPs (Floating Point Operations) 
required to process one input item (image 
or text fragment). 
We collect inference FLOPs for many different DNNs architectures 
following a comprehensive literature review. 
Since hardware manufacturers have been working on specific chips for DNN, adapting the hardware to a specific case of use leads to performance and efficiency improvements. We collect hardware data over the recent years, and estimate how many FLOPs can be obtained with one Joule with each chip. Having all this data we  finally estimate how much energy is needed to perform one inference step with a given DNN. Our main objective is to study the evolution of the required energy for one prediction over the years. 


The main findings and contributions of this paper are to (1) showcase that better results for DNN models are in part attributable to algorithmic improvements and not only to more computing power; (2) determine how much hardware improvements and specialisation is decreasing DNNs energy consumption; (3) report that, while  energy consumption is still increasing exponentially for new cutting-edge models, DNN inference energy consumption could be maintained low for increasing performance if the efficient models that come relatively soon after the breakthrough are selected.  
%
%

We provide all collected data and performed estimations as a data set, publicly available in the appendixes and as a GitHub repository\footnote{Temporary copy in: \url{https://bit.ly/3DTHvFC}}
. The rest of the paper covers the background, introduces the methodology and presents the analysis of hardware and energy consumption of DNN models and expounds on some forecasts. Discussion and future work close the paper.


\section{Background}\label{section:background}

In line with other areas of computer science, there is some previous work that analyses compute and its cost for AI, and DNNs more specifically. Recently, OpenAI carried out a detailed analysis about AI efficiency \citep{hernandez2020measuring}, focusing on the amount of compute used to train models with the ImageNet dataset. 
They show that 44 times less compute was required in 2020 to train a network with the performance AlexNet achieved seven years before. 

However, a demand for better task performance, 
linked with more complex DNNs and larger volumes of data to be processed, the growth in demand for AI compute is still growing fast. 
\citep{thompson2020computational} reports the computational demands of several Deep Learning applications, showing that progress in them is strongly reliant on increases in computing power. 
AI models have doubled the computational power used every 3.4 months since 2012 \citep{ai_and_compute}. The study \citep{gholami2020ai_and_memory_wall} declare similar scaling rates for AI training compute to \citep{ai_and_compute} and they forecast that DNNs memory requirements will soon become a problem.
This exponential trend seems to impose a limit on how far we can improve performance in the future without a paradigm change. 

Compared to training costs, there are fewer studies on inference costs, despite using a far more representative share of compute and energy. \citeauthor{canziani2017analysis} (\citeyear{canziani2017analysis}) study accuracy, memory footprint, parameters, operations count, inference time and power consumption of 14 ImageNet models. To measure the power consumption they execute the DNNs on a  NVIDIA Jetson TX1 board. 
A similar study \citep{7723730} measures energy efficiency, Joules per image, for a single forward and backward propagation iteration (a training step). This study benchmarks 4 Convolutional Neural Networks (CNNs) on CPUs and GPUs on different frameworks. Their work shows that GPUs are more efficient than CPUs for the CNNs analysed. Both publications analyse model efficiency, but they do this for very concrete cases. We 
analyse a greater number of DNNs and hardware components in a longer time frame.

These and other papers are key in helping society and AI researchers realise the issues about efficiency and energy consumption. \citeauthor{strubell2019energy} (\citeyear{strubell2019energy}) estimate the energy consumption, the cost and CO\textsubscript{2} emissions of training various of the most popular NLP models.
\citeauthor{henderson2020towards} (\citeyear{henderson2020towards}) performs a systematic reporting of the energy and carbon footprints of reinforcement learning algorithms. 
\citeauthor{bommasani2021opportunities} (\citeyear{bommasani2021opportunities}) (section 5.3) seek to identify assumptions that shape the calculus of environmental impact for foundation models. 
\citeauthor{schwartz2019green} (\citeyear{schwartz2019green}) analyse training costs and propose that researchers should put more attention on efficiency and they should report always the number of FLOPs.  %
These studies contribute to a better assessment of the problem and more incentives for their solution. For instance, 
new algorithms and architectures such as EfficientNet \citep{tan2020efficientnet} and EfficientNetV2 \citep{tan2021efficientnetv2} have aimed at this reduction in compute.  

When dealing about computing effort and computing speed (hardware performance), terminology is usually confusing. For instance, the term `compute' is used ambiguously, sometimes applied to the number of operations or the number of operations per second. However, it is important to clarify what kind of operations and the acronyms for them. In this regard, we will use the acronym FLOPS to measure hardware performance, by referring to the number of floating point operations {\em per second}, as standardised in the industry, while FLOPs will be applied to the amount of computation for a given task (e.g., a prediction or inference pass), by referring to the number of operations, counting a multiply-add operation pair as two operations. An extended discussion about this can be found in the appendix. 

\section{Methodology}\label{section:methodology}

%
We collect most of our information directly from research papers that report results, compute and other data for one or more newly introduced techniques for the benchmarks and metrics we cover in this work. We manually read and inspected the original paper and frequently explored the official GitHub repository, if exists. However, often there is missing information in these sources, so we need to get the data from other sources, namely:
\begin{itemize}
\item \emph{Related papers}: usually the authors of another paper that introduces a new model compare it with previously existing models, providing further information.
\item \emph{Model implementations}: {\small \ttfamily{PyTorch}}  \citep{pytroch} 
contains many (pre-trained) models, and their performance is reported. Other projects do the same (see, e.g., \citep{pretrainedmodels,cvpytroch}).
\item \emph{Existing data compilations}: there are some projects and public databases collecting information about deep learning architectures and their benchmarks, e.g., \citep{git_convnet,coleman2017dawnbench,mattson2020mlperf,git_nlp,pwc}.   
\item \emph{Measuring tools}: when no other source was available or reliable, 
we used the {\small \ttfamily ptflops} library \citep{ptflops} or similar tools 
to calculate the model's FLOPs and parameters (when the implementation is available). 
\end{itemize}

\noindent Given this general methodology, we now discuss in more detail how we made the selection of  CV and NLP models, and the information about hardware. 

\subsection{CV Models Data Compilation}

There is a huge number of models for image classification, so we selected models based on two criteria: popularity and accuracy. For  popularity we looked at the times that the paper presenting the model is cited on Google Scholar and 
whether the model appears mentioned in other papers (e.g., for comparative analyses). 
We focused on model's accuracy as well because having the best models per year in terms of accuracy is necessary for analysing progress. 
 To achieve this we used existing compilations \citep{pwc} and filtered by year and accuracy. For our selection, accuracy was more important than popularity for recent models, 
 as they are less cited than the older ones because they have been published for a shorter time. 
Once we selected the sources for image classification models, we collected the following information: Top-1 accuracy on ImageNet, number of parameters, FLOPs per forward pass, release date and training dataset. Further details about model selection, FLOPs estimation, image cropping \citep{krizhevsky2012ImageNet}  and resolution \citep{simonyan2015deep,zhai2021scaling} can be found in the Appendix (and Table \ref{tab:modelDetails}).

\subsection{NLP Models Data Compilation}

For NLP models we noted that there is much less information about inference  (e.g., FLOPs) and the number of models for which we can get the required information is smaller than for CV. We chose GLUE for being sufficiently representative  and its value determined for a good number of architectures. 
To keep the numbers high we 
just included all the models since 2017 for which we found inference compute estimation \citep{clark2020electra}. Further details about FLOPs estimation and counting can be found in the Appendix (selected models in in Table \ref{tab:nlp_data}). 


\subsection{Hardware Data Compilation}

Regarding hardware evolution, we collected data for Nvidia GPUs\footnote{\url{https://developer.nvidia.com/deep-learning}}. We chose Nvidia GPUs because they represent one of the most efficient hardware platforms for DNN\footnote{We considered Google's TPUs (\url{https://cloud.google.com/tpu?hl=en}) for the analysis but there is not enough public information about them, as they are not sold but only available as a service.} and they have been used for Deep Learning in the last 10 years, so we have a good temporal window for exploration. 
In particular, we collected GPU data for Nvidia GPUs from 2010 to 2021. The collected data is: FLOPS, memory size, power consumption (reported as Thermal Design Power, TDP) and launch date. As explained before, FLOPS is a measure of computer performance. From the FLOPS and power consumption we calculate the efficiency, dividing FLOPS by Watts. 
We use TDP and the reported peak FLOPS to calculate efficiency. 
This means we are considering the efficiency (GLOPS/Watt) when the GPU is at full utilisation. In practice the efficiency may vary depending on the workload, but we consider this estimate (``peak FLOPS''/TDP) accurate enough for analysing the trends and for giving an approximation of energy consumption. In our compilation there are desktop GPUs and server GPUs. We pay special attention to server GPUs released in the last years, because they are more common for AI, and DNNs in particular. A discussion about discrepancies between theoretical and real FLOPS as well as issues regarding Floating Point (FP) precision operations can be found in the Appendix.   

\section{Computer Vision 
Analysis}\label{section:CV}

In this section, we analyse the evolution of ImageNet \citep{deng2009ImageNet} (one pass inference) according to performance and compute. Further details  in the Appendix.

\subsection{Number of Parameters and FLOPs}\label{section:params}

The number of parameters is usually reported, but it is not directly proportional to compute. For instance, in CNNs, convolution operations dominate the computation: if $d$, $w$ and $r$ represent the network's depth, widith and input resolution, the FLOPs grow following the relation \citep{tan2020efficientnet}:
$$\mbox{FLOPs} \propto d + w^2 + r^2$$
\noindent
This means that FLOPs do not directly  depend on the number of parameters. Parameters affect network depth ($d$) or width ($w$), but distributing the same number of parameters in different ways will result in different numbers of FLOPs. Moreover, the resolution ($r$) does not depend on the number of parameters directly, because the input resolution can be increased without increasing network size. 

\begin{figure}[!h]
\centering
\includegraphics[width=0.9\columnwidth]{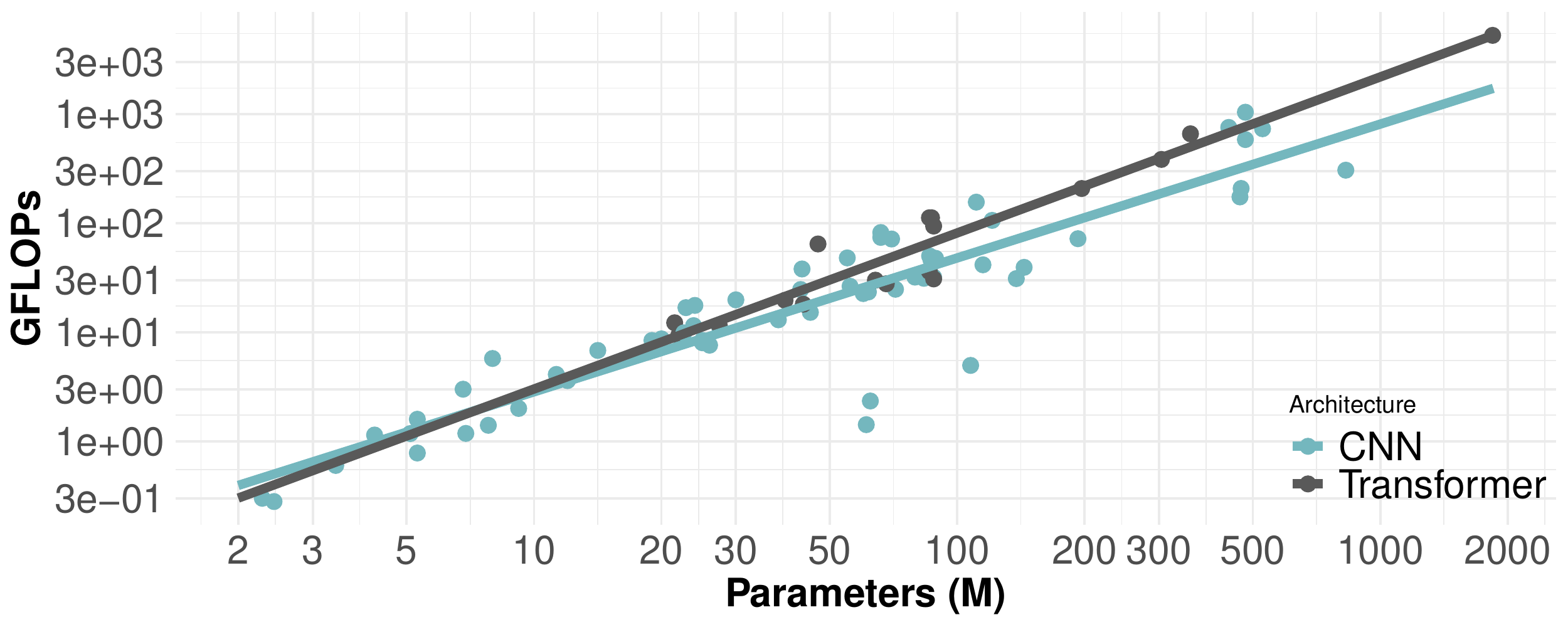}
\caption{\label{fig:params}Relation between the number of parameters and FLOPs (both axes are logarithmic).  
}
\end{figure}

Despite this, Fig.~\ref{fig:params} shows a linear relation between FLOPs and parameters. We attribute this to the balanced scaling of $w$, $d$ and $r$. These dimensions are usually scaled together with bigger CNNs using higher resolution. Note that recent transformer models \citep{vaswani2017attention} do not follow the growth relation presented above. However, the correlation between the number of parameters and FLOPs for CNNs is 0.772 and the correlation for transformers is 0.994 (Fig.~\ref{fig:params}). This suggests that usually in both architectures parameters and FLOPs scale in tandem.
%
We will use
FLOPs, as they allow us to estimate the needed energy relating hardware FLOPS with required FLOPs for a model \citep{how_fast,clark2020electra}. 

\subsection{Performance and Compute} 

There has been very significant progress 
for ImageNet. In 2012, 
AlexNet 
achieved 56\% Top-1 accuracy (single model, one crop). In 2021, Meta Pseudo Labels  (EfficientNet-L2) 
achieved 90.2\% Top-1 accuracy (single model, one crop). However, this increase in accuracy comes with an increase in the required FLOPs for a forward pass. A forward pass for AlexNet 
is 
1.42 GFLOPs 
while for EfficientNet-L2 is 1040 GFLOPs (details in the appendix). 

\begin{figure}[!h]
\centering
\includegraphics[width=0.9\columnwidth]{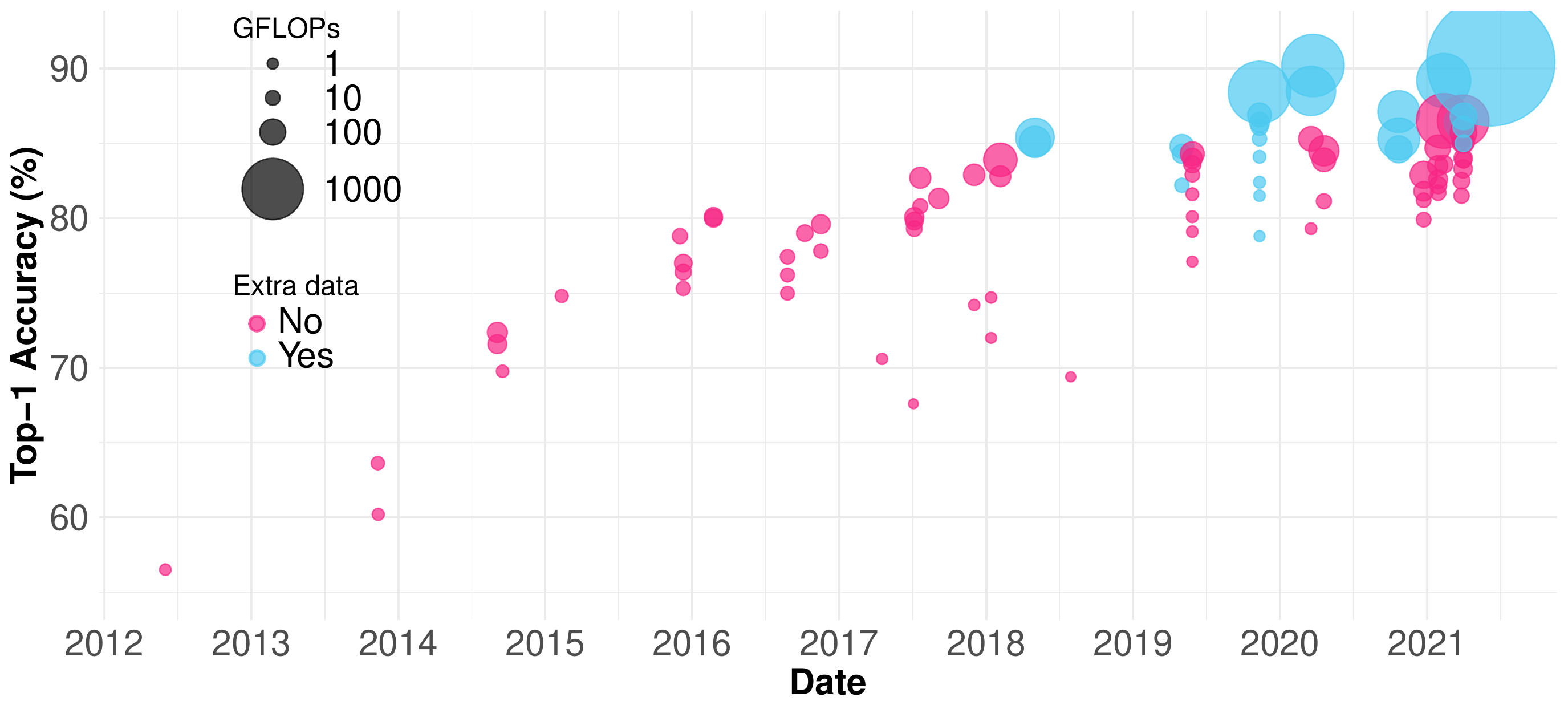}
\caption{\label{fig:acc_years}Accuracy evolution over the years. The size of the balls represent the GFLOPs of one forward pass. }
\end{figure}

Fig.~\ref{fig:acc_years} shows the evolution from 2012 to 2021 in ImageNet accuracy (with the size of the bubbles representing the FLOPs of one forward pass). In recent papers some researchers began using more data than those available in ImageNet1k for training the models. 
However, using extra data only affects training FLOPs, but does not affect the computational cost for {\em inferring} each classification (forward pass). 

\begin{figure}[!h]
\centering
\includegraphics[width=0.9\columnwidth]{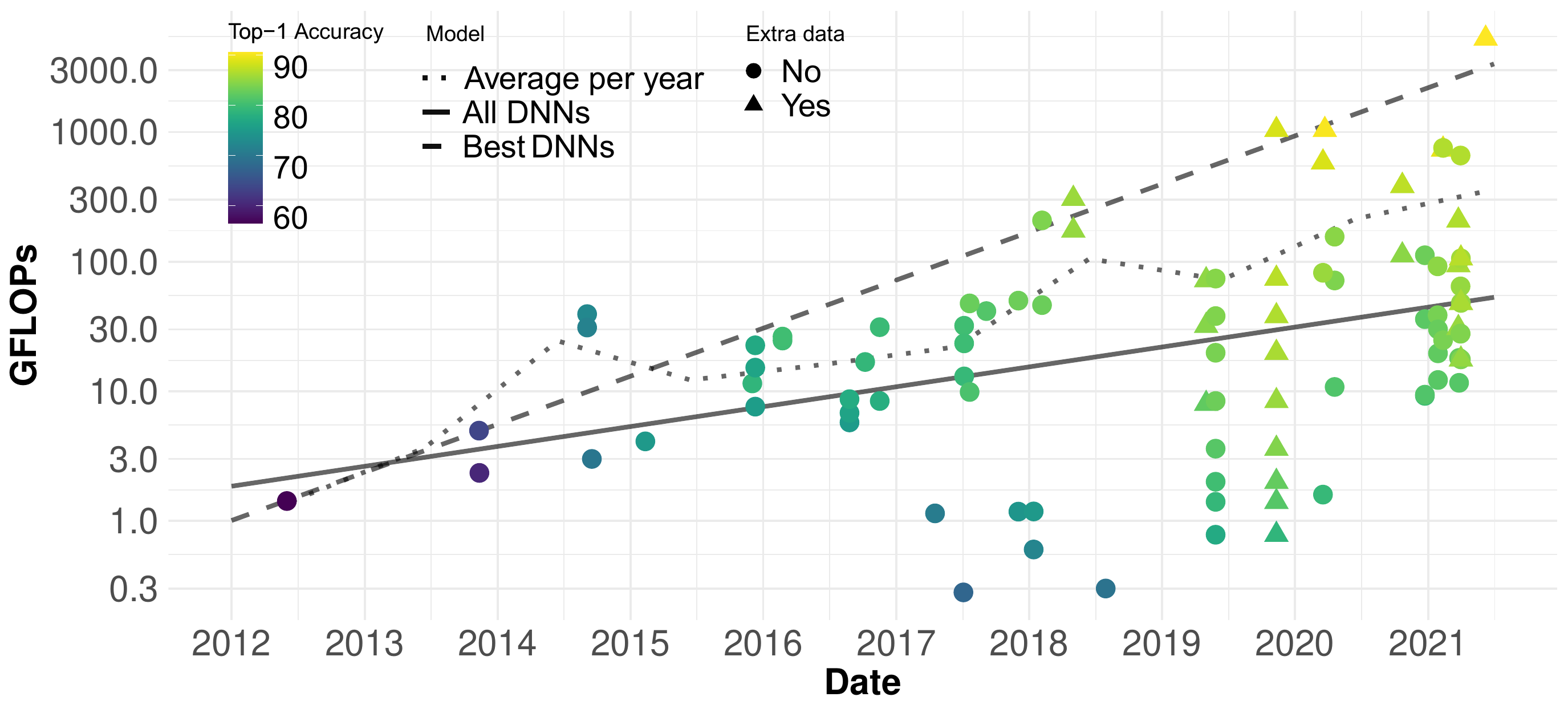}
\caption{\label{fig:date_flops}GFLOPs over the years. The dashed line is a linear fit (note the logarithmic \yaxis) for the models with highest accuracy per year. The solid line includes all points.}
\end{figure}

If we only look at models with the best accuracy for each year we can see an exponential growth in compute (measured in FLOPs). This can be observed clearly in Fig.~\ref{fig:date_flops}: the dashed line represents an exponential growth (shown as a linear fit since the \yaxis is logarithmic). The line is fitted with the models with highest accuracy for each year. However not all models released in the latest years need so much compute. 
This is reflected by the solid line, which includes all points.
We also see that for the same number of FLOPs we have models with increasing accuracy as time goes by.

In  Table \ref{table:tableAlexNet} there is a list of models having similar number of FLOPs as AlexNet. In 2019 we have a model (EfficientNet-B1) with the same number of operations as AlexNet achieving a Top-1 accuracy of 79.1\% without using extra data, and a model (NoisyStudent-B1) achieving Top-1 accuracy of 81.5\% using extra data. In a period of 7 years, we have models with similar computation  with much higher accuracy. We observe that when a SOTA model is released it usually has a huge number of FLOPs, and therefore consumes a large amount of energy, but in a couple of years there is a model with similar accuracy but with much lower number of FLOPs. These models are usually those that become popular in many industry applications. This observation confirms that better results for DNN models of {\em general use} are in part attributable to algorithmic improvements and not only to the use of more computing power. 

\begin{table}[!h]
\centering
\resizebox{0.7\columnwidth}{!}{%
\begin{tabular}{@{}lccc@{}}
\toprule
\textbf{Model} & \textbf{Top-1 Accuracy} & \textbf{GFLOPs} & \textbf{Year} \\ \midrule
\textbf{AlexNet} \citep{krizhevsky2012ImageNet} & 56.52 & 1.42 & 2012 \\
\textbf{ZFNet} \citep{zeiler2013visualizing} & 60.21 & 2.34 & 2013 \\
\textbf{GoogleLeNet} \citep{szegedy2014going} & 69.77 & 3.00 & 2014 \\
\textbf{MobileNet} \citep{howard2017mobilenets} & 70.6 & 1.14 & 2017 \\
\textbf{MobileNetV2 1.4} \citep{sandler2019mobilenetv2} & 74.7 & 1.18 & 2018 \\
\textbf{EfficientNet-B1} \citep{tan2020efficientnet}& 79.1 & 1.40 & 2019 \\
\textbf{NoisyStudent-B1} \citep{xie2020selftraining} & 81.5 & 1.40 & 2019 \\ \bottomrule
\end{tabular}%
}
\vspace{0.3cm}
\caption{Results for several DNNs with a similar number of FLOPs as AlexNet. \label{table:tableAlexNet}}
\end{table}

\begin{figure}[!h]
\centering
\includegraphics[width=0.9\columnwidth]{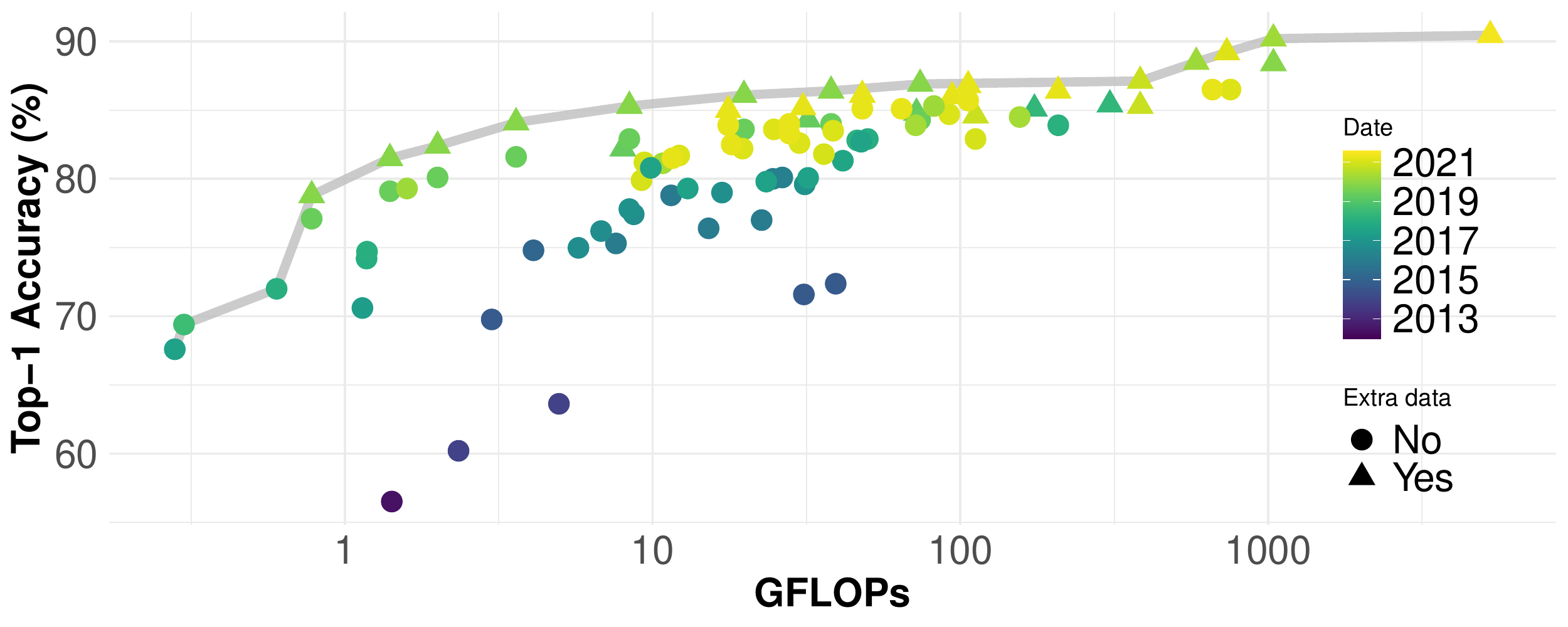}
\caption{\label{fig:acc_flops}Relation between accuracy and GFLOPs.}
\end{figure}

Finally, Fig.~\ref{fig:acc_flops} shows that the Pareto frontier (in grey) 
is composed of new models (in yellow and green), whereas old models (in purple and dark blue) are relegated below the Pareto. 
As expected, the models which use extra data are normally those forming the Pareto frontier. Let us note again that extra training data does not affect inference GFLOPs.

\section{Natural Language 
Analysis}\label{section:NLP}

In this section, we analyse the trends in performance and inference compute for NLP models. 
To analyse performance we use GLUE, which is a popular benchmark for natural language understanding, one key task in NLP. The GLUE benchmark\footnote{Many recent models are evaluated on SUPERGLUE, but we choose GLUE to have a temporal window for our analysis.} is composed of nine sentence understanding tasks, which cover a broad range of domains. The description of each task can be found in \citep{wang2019glue}.

\subsection{Performance and Compute}


We represent the improvement on the GLUE score in relation to GFLOPs over the years 
in  Fig.~\ref{fig:flops_glue} (and in Fig.~\ref{fig:date_glue} in the Appendix). GFLOPs are for single input of length 128, which is a reasonable sequence length for many use cases, being able to fit text messages or short emails. We can observe a very similar evolution to the evolution observed in ImageNet: SOTA models require a large number of FLOPs, but in a short period of time other models appear, which require much fewer FLOPs to reach the same score. 
There are many models that focus on being efficient instead of reaching  high score, and this is reflected in their names too (e.g., MobileBERT \citep{sun2020mobilebert} and SqueezeBERT \citep{iandola2020squeezebert}). We note that the old models become inefficient (they have lower score with higher number of GLOPs) compared to the new ones, as it happens in CV models. 



\begin{figure}[!h]
\centering
\includegraphics[width=0.9\columnwidth]{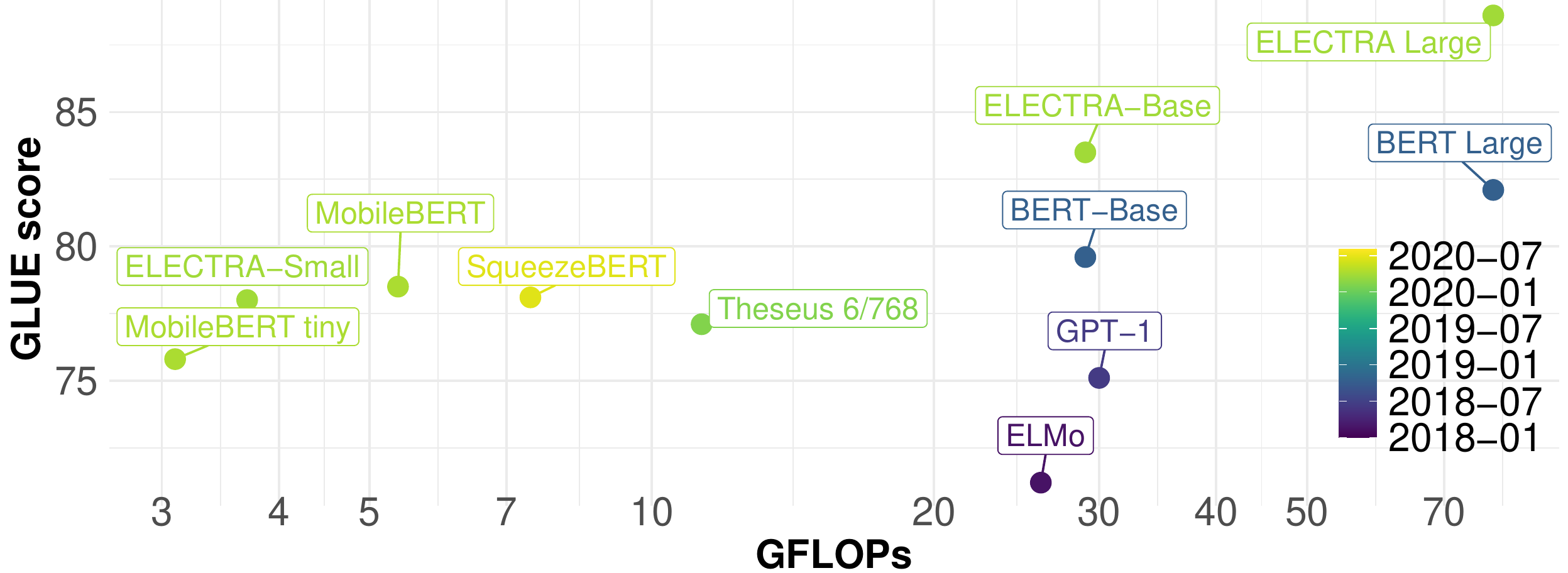}
\caption{\label{fig:flops_glue}GFLOPs per token analysis for NLP models.}
\end{figure}

\subsection{Compute Trend}

\begin{figure}[!h]
\centering
\includegraphics[width=0.9\columnwidth]{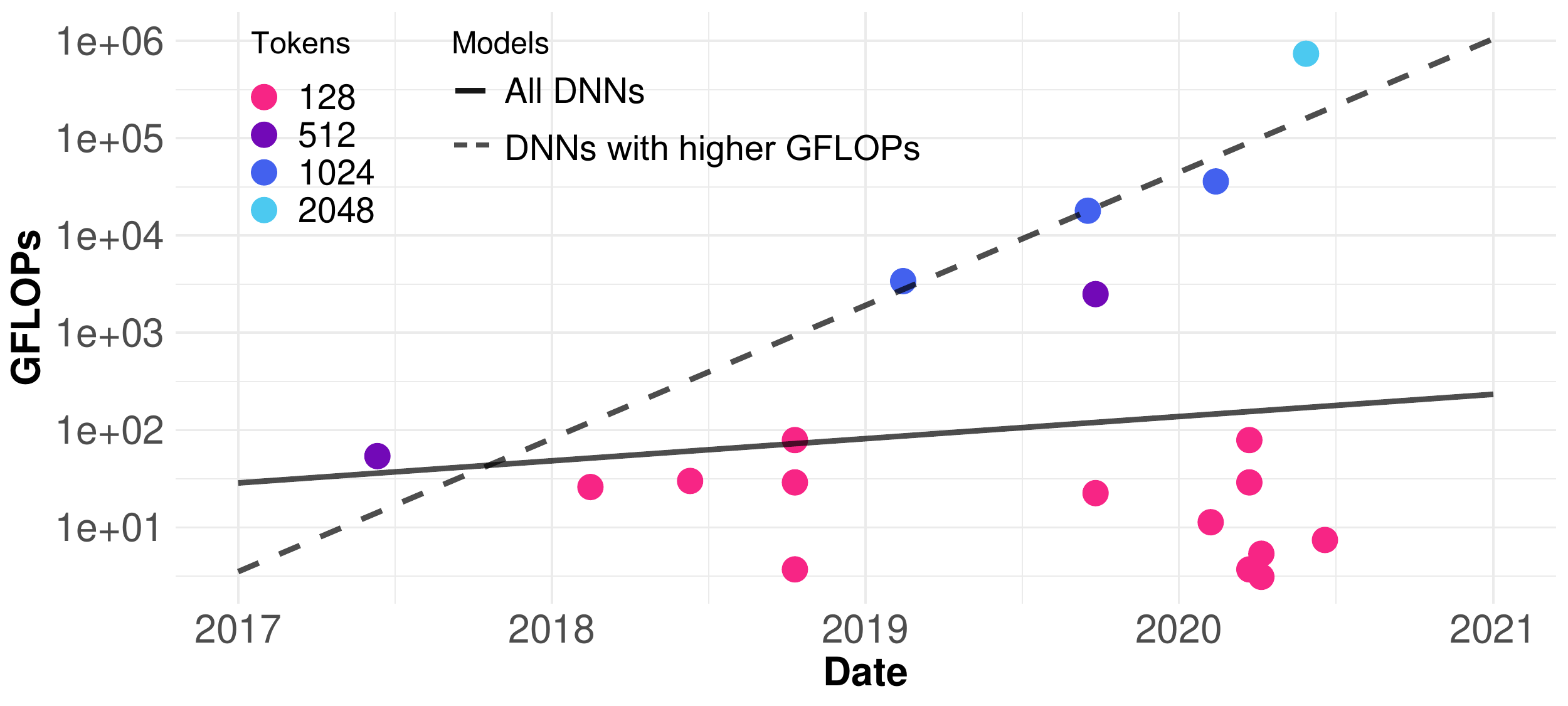}
\caption{\label{fig:date_flops_nlp}GFLOPs per token analysis for NLP models.}
\end{figure}

In Fig.~\ref{fig:date_flops_nlp} 
we include all models (regardless of having performance results) for which we found inference FLOPs estimation. The dashed line 
adjusts to the models with higher GFLOPs (models that, when released, become the most demanding model) and the solid line 
to 
all NLP models. In this plot we indicate the input sequence length, because in this plot we represent models with different input sequence lengths. We observe a similar trend as in CV: the GFLOPS of the most cutting-edge models have a clear exponential growth, while the general trend, i.e., considering all models, does not scale so aggressively. Actually, there is a good pocket of low-compute models in the last year.

\section{Hardware Progress}\label{section:hardware}

\begin{figure}[!h]
\centering
\includegraphics[width=0.9\columnwidth]{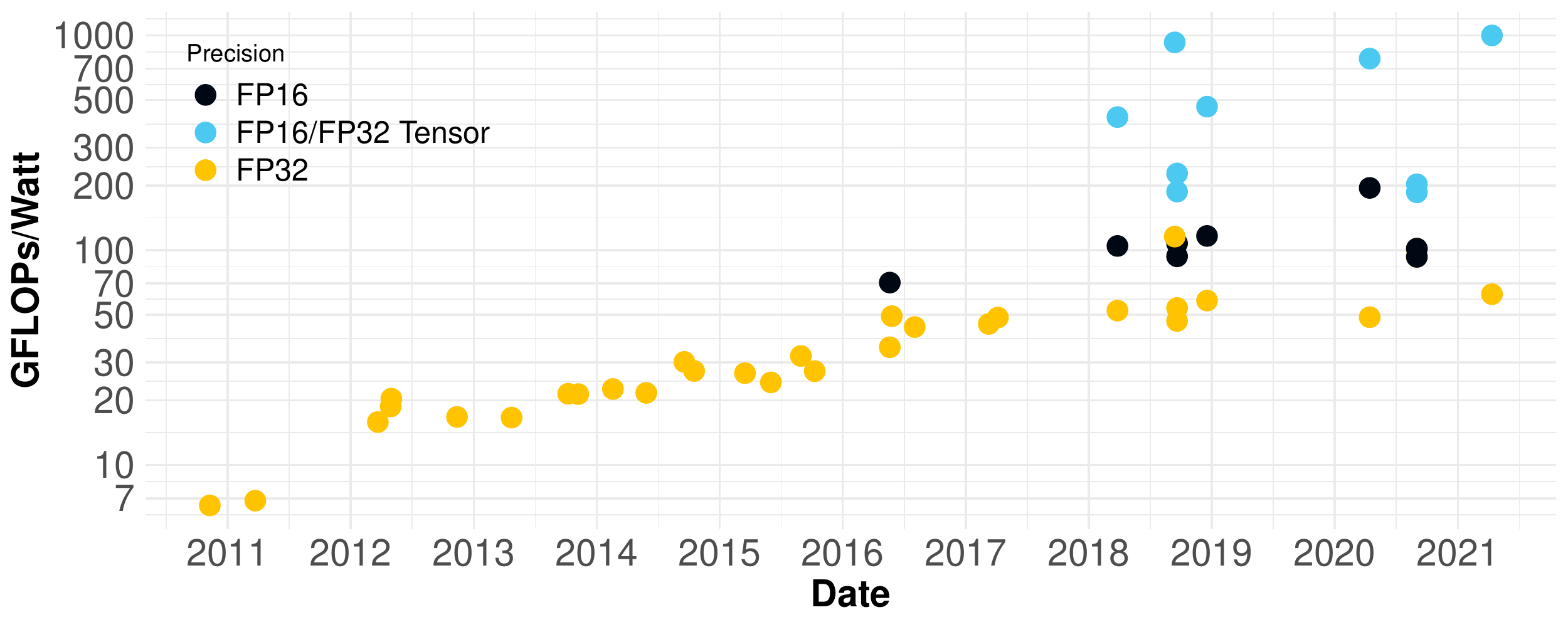}
\caption{\label{fig:gpus}Theoretical Nvidia GPUs GFLOPS per Watt. Data in Table \ref{tab:gpu_theorical} in the appendix.}
\end{figure}

We use FLOPS as a measure of hardware performance and FLOPS/Watt as a measure of hardware efficiency. We collected performance for different precision formats and tensor cores for a wide range of GPUs. The results are shown in Fig.~\ref{fig:gpus}. Note that the \yaxis is in logarithmic scale. 
Theoretical FLOPS for tensor cores are very high in the plot. However, the actual performance for inference using tensor cores is not so high, if we follow a more realistic estimation for the Nvidia GPUs (V100, A100 and T4\footnote{Specifications in: \url{https://www.nvidia.com/en-us/data-center/}.}).  
 The details of this estimation 
 are shown in Table \ref{tab:gpu_speed_up_models} in the appendix. 

\begin{figure}[!h]
\centering
\includegraphics[width=0.9\columnwidth]{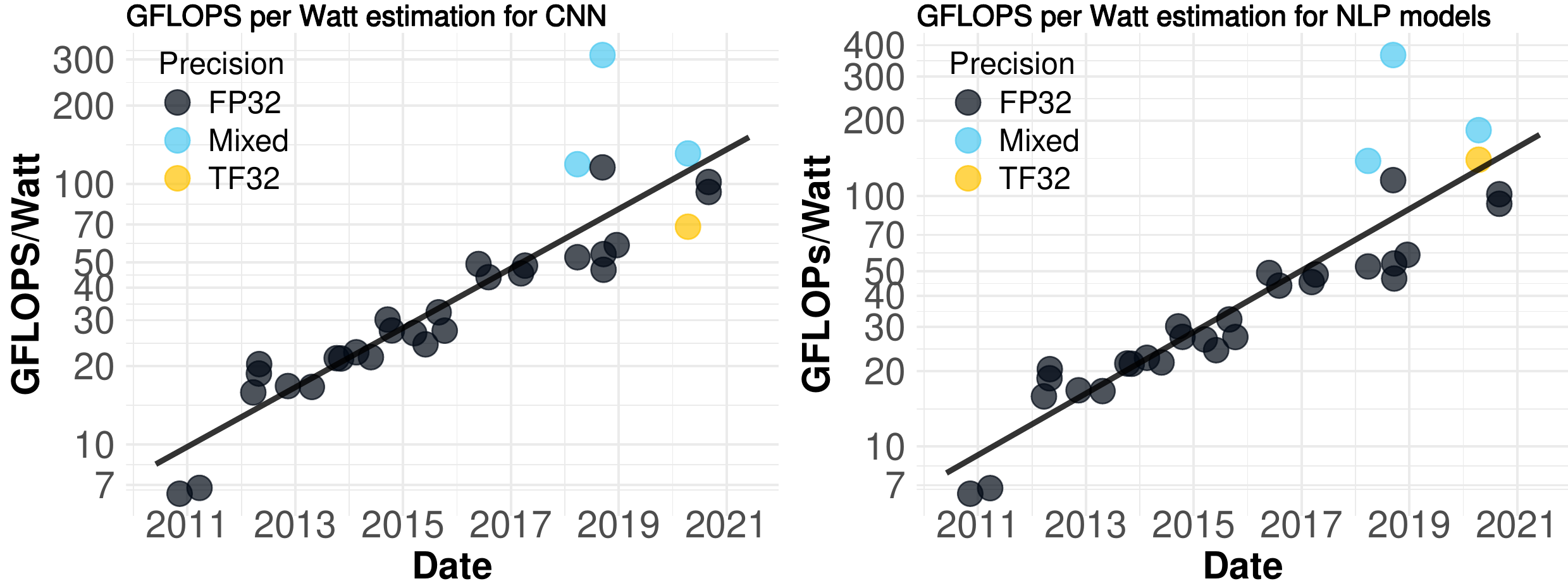}
\caption{\label{fig:gpus_adapted}Nvidia GPU GFLOPS per Watt adapted for CV (CNNs) and NLP models. Data in Table \ref{tab:gpu_models} in the appendix.} 
\end{figure}

With these estimations we obtained good linear fits (with the \yaxis in logarithmic scale) to each data set, one for CV and another for NLP, as shown by the solid lines in Fig.~\ref{fig:gpus_adapted}. Notice that there is a particular point in Fig.~\ref{fig:gpus_adapted} for year 2018 that stands out among the others by a large margin. This corresponds to T4 using mixed precision, a GPU specifically designed for inference, and this is the reason why it is so efficient for this task. 

\section{Energy Consumption Analysis}

Once we have estimated the inference FLOPs for a range of models and the GFLOPS per Watt for different GPUs, we can estimate the energy (in Joules) consumed in one inference. We do this by dividing the FLOPs for the model by FLOPS per Watt for the GPU. But how can we choose the FLOPS per Watt that correspond to the model? 
We use the models presented in Fig.~\ref{fig:gpus_adapted} to obtain an estimation of GLOPS per Watt {\em for the model's release date}. In this regard, \citeauthor{henderson2020towards} (\citeyear{henderson2020towards}) report that FLOPs for DNNs can be misleading sometimes, due to underlying optimisations at the firmware, frameworks, memory and hardware that can influence energy efficiency. They show that energy and FLOPs are highly correlated for the same architecture, but the correlation decreases when different architectures are mixed.  We consider that this low correlation does not affect our estimations significantly as we analyse the trends through the years and we fit in the exponential scale, where dispersion is reduced. To perform a more precise analysis it would be necessary to measure power consumption for each network with the original hardware and software, as unfortunately 
the required energy per (one) inference is rarely reported. 

\begin{figure}[!h]
\centering
\includegraphics[width=0.9\columnwidth]{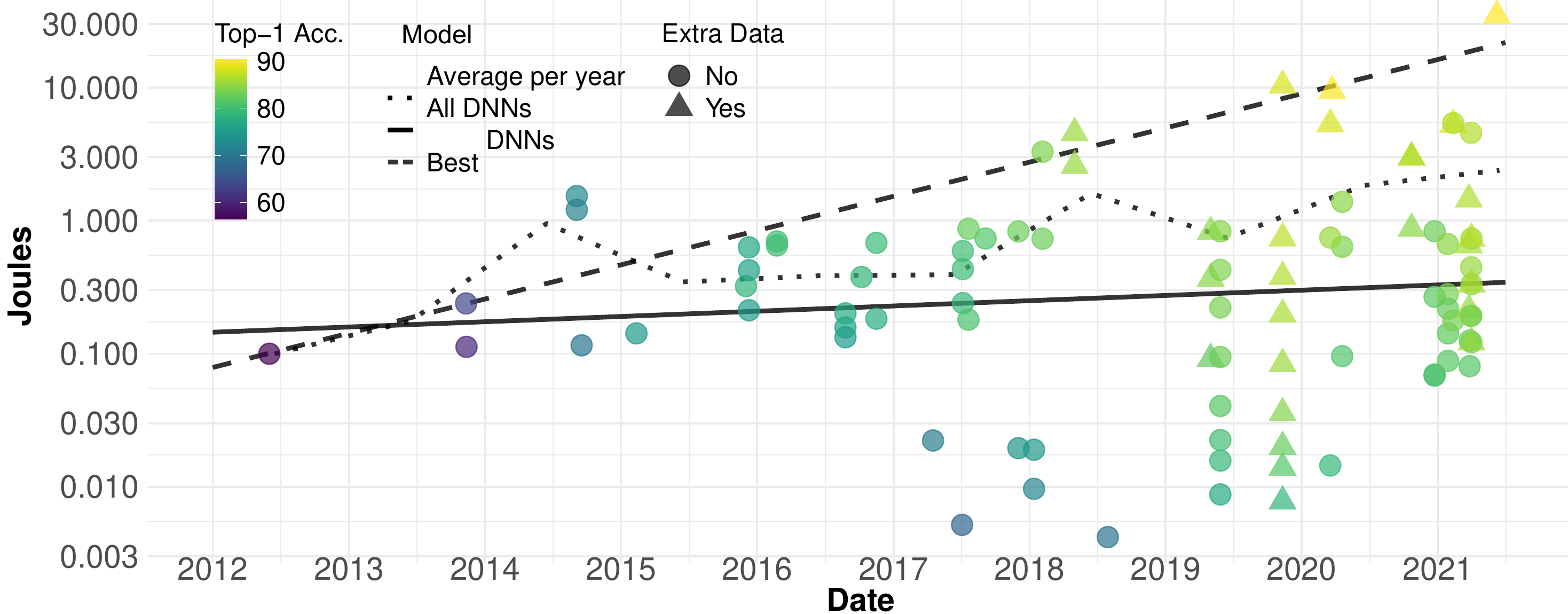}
\caption{\label{fig:watts_estimation}Estimated Joules of a forward pass (CV). The dashed line is a linear fit (logarithmic \yaxis) for the models with highest accuracy per year. The solid line fits all models.}
\end{figure}

\begin{figure}[!h]
\centering
\includegraphics[width=0.9\columnwidth]{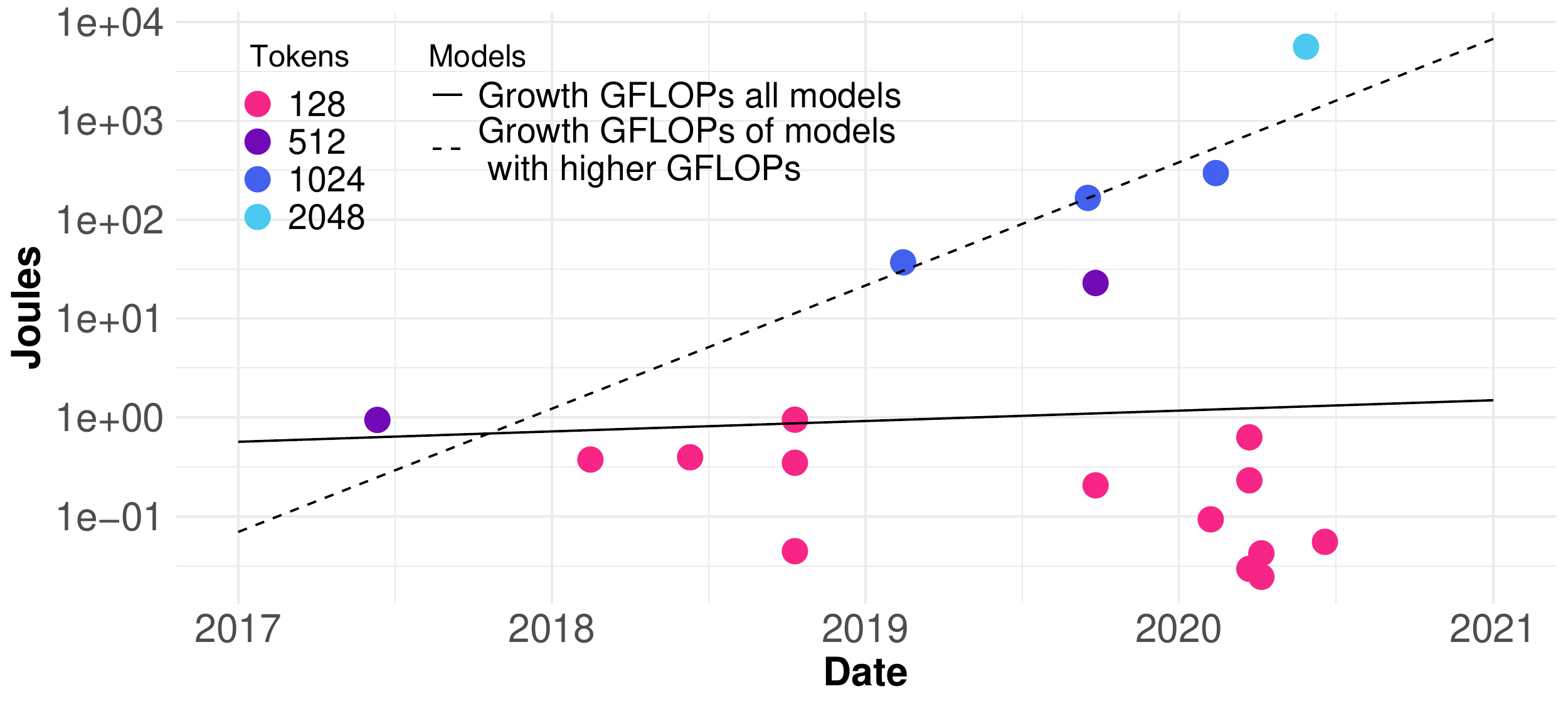}
\vspace{-0.2cm}
\caption{\label{fig:watts_estimation_nlp}Estimated Joules of a forward pass (NLP). Same interpretation as in Fig.~\ref{fig:watts_estimation}.
}
\end{figure}

We can express the efficiency metric FLOPS per Watt as FLOPs per Joule, as shown in Eq. \ref{eq:FLOPS/Watt}. Having this equivalence we can use it to divide the FLOPs needed for a forward pass and obtain the required Joules, see Eq. \ref{eq:joules}.  Doing this operation we obtain the consumed energy in Joules. 

\begin{equation}
\label{eq:FLOPS/Watt}
\mbox{
\mbox{Efficiency} $= \frac{\mbox{HW Perf.}}{\mbox{Power}} \:\: \mbox{in units:} \:\:
\cfrac{FLOPS}{Watt} = \cfrac{FLOPs/s}{Joules/s} = \cfrac{FLOPs}{Joule}$}
\end{equation}
\vspace{-0.1cm}
\begin{equation}
\label{eq:joules}
\mbox{
\mbox{Energy} $= \frac{\mbox{Fwd. Pass}}{\mbox{Efficiency}} \:\: \mbox{in units:} \:\:
\cfrac{FLOPs}{FLOPs/Joule} =  Joule
$}
\end{equation}

Applying this calculation to all collected models we obtain Fig.~\ref{fig:watts_estimation} for CV. The dashed line represents an exponential trend (a linear fit as the \yaxis is logarithmic), adjusted to the models with highest accuracy for each year, like in Fig.~\ref{fig:acc_years}, and the dotted line represent the average Joules for each year. By comparing both plots we can see that hardware progress softens the growth observed for FLOPs, 
but the growth is still clearly exponential for the models with high accuracy. The solid line is almost horizontal, but in a logarithmic scale this may be interpreted as having an exponential growth with a small base or a linear fit on the semi log plot that is affected by the extreme points. 
In Fig.~\ref{fig:watts_estimation_nlp}  we do the same for NLP models and we see a similar picture.

\begin{figure}[!h]
\centering
\includegraphics[width=0.9\columnwidth]{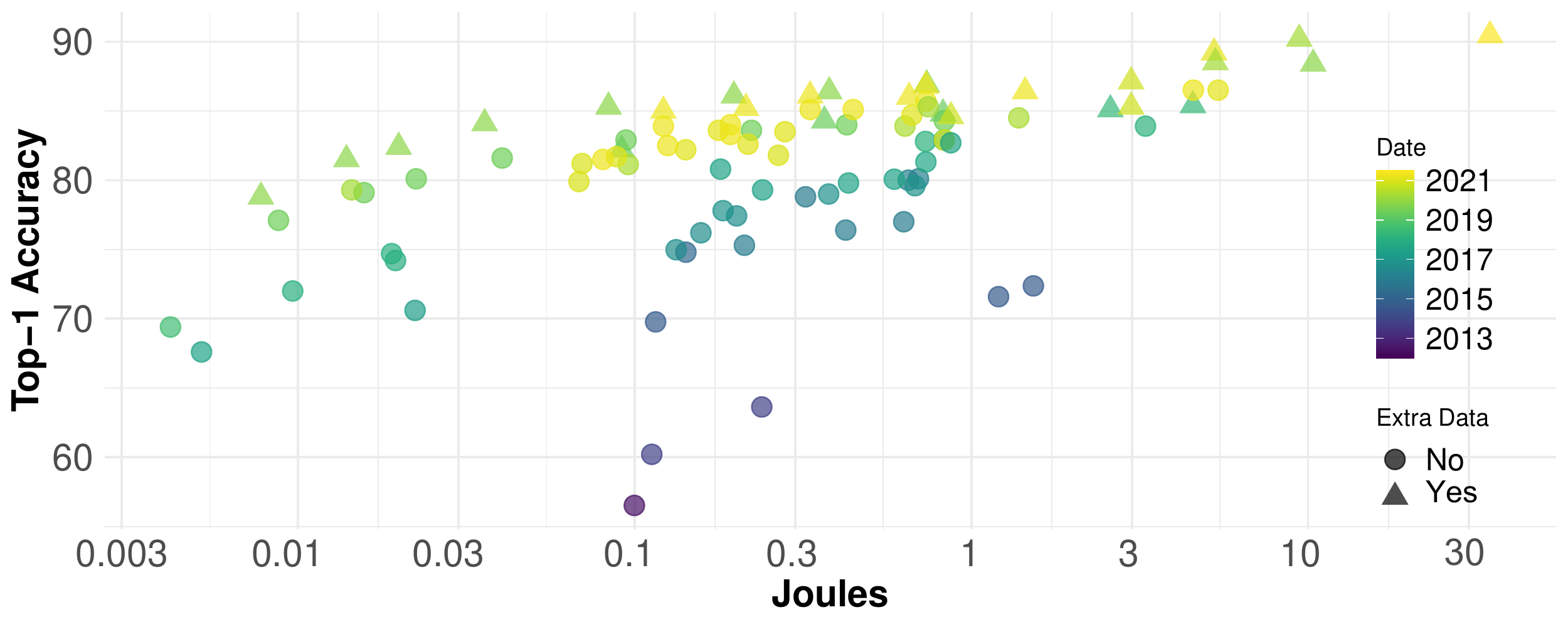}
\caption{\label{fig:joules_acc}Relation between Joules and Top-1 Accuracy over the years (CV, ImageNet).}
\end{figure}

\begin{figure}[!h]
\centering
\includegraphics[width=0.9\columnwidth]{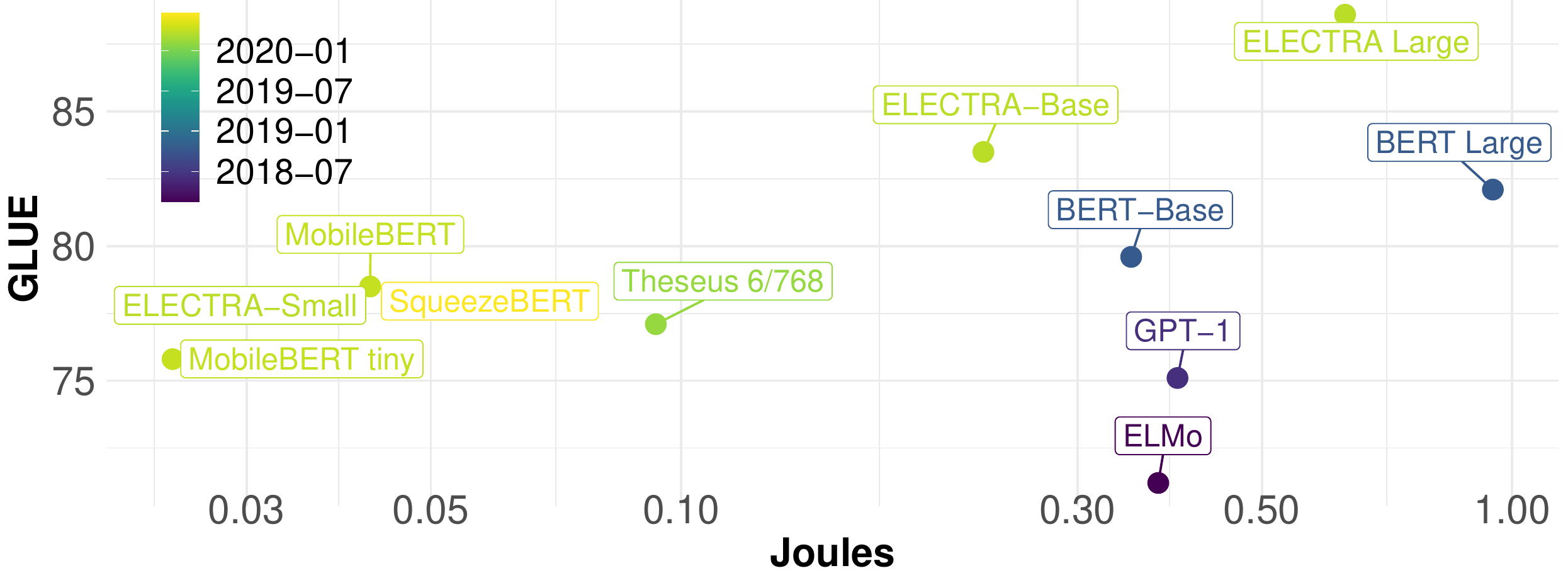}
\caption{\label{fig:joules_acc_nlp}Relation between Joules and GLUE score over the years (NLP, GLUE).
}
\end{figure}

Fig.~\ref{fig:joules_acc} shows the relation between Top-1 Accuracy and Joules. Joules are calculated in the same way as in Fig.~\ref{fig:watts_estimation}. The relation is similar as the observed in Fig.~\ref{fig:acc_flops}, but in Fig.~\ref{fig:joules_acc} the older models are not only positioned further down in the \yaxis (performance) but they tend to cluster on the bottom right part of the plot (high Joules), so their position on the \yaxis is worse than for  
Fig.~\ref{fig:acc_flops} due to the evolution in hardware. This is even more clear for NLP, as seen in Fig.~\ref{fig:joules_acc_nlp}.

\section{Forecasting and Multiplicative Effect}\label{section:forecast}

In our analysis we see that DNNs as well as hardware are improving their efficiency and do not show symptoms of standstill. This is consistent with most studies in the literature: performance will continue growing as compute grows, but at the same time efficiency is increasing. However, this is the first work that analyses whether these two things cancel, especially when we analyse inference and not training. Our conclusion is that they not cancel out for the cutting-edge models of each moment but this is less clear for the regular models in {\em general use} by industries and invididuals. 

However, since we are focusing on inference costs, we need to consider the multiplicative factor. How many inferences are performed {\em per capita}? This has definitely increased very significantly with the use of smart devices, Internet of things and many other devices around us, which are incorporating DNN-based services. However, how many inference passes per capita do we have at this moment, and how is this growing? This is very difficult to estimate, and we leave it for future work. However, it is interesting to a\-na\-ly\-se possible hypotheses: assume there is one inference pass of a neural network application per second per capita. What would this imply in terms of energy consumption?

\begin{figure}[!t]
\centering
\includegraphics[width=0.9\columnwidth]{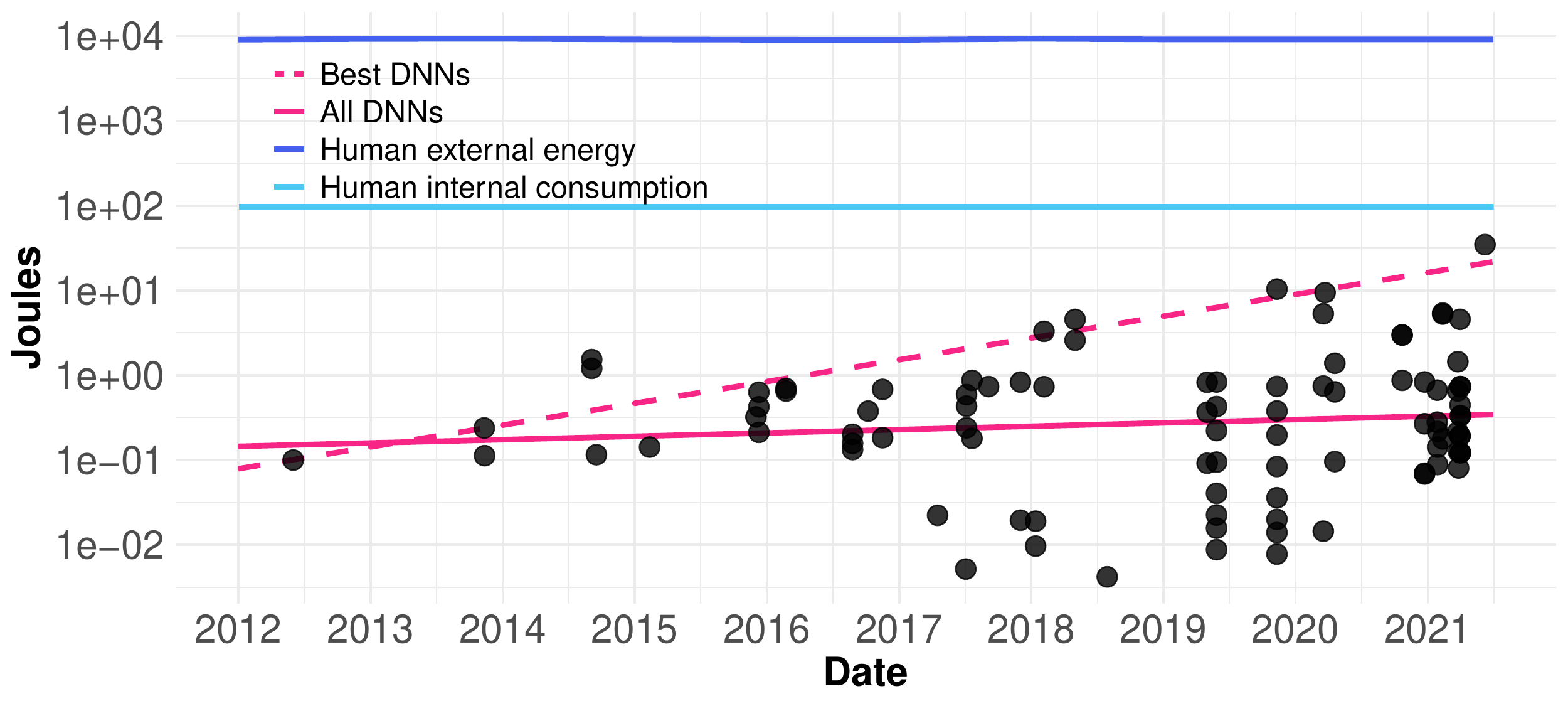}
\caption{\label{fig:joules_year_human}Estimated Joules per forward pass (e.g., one prediction) compared to human energy consumption in 1s (CV).}
\end{figure}

\begin{figure}[!t]
\vspace{-0.3cm}
\centering
\includegraphics[width=0.9\columnwidth]{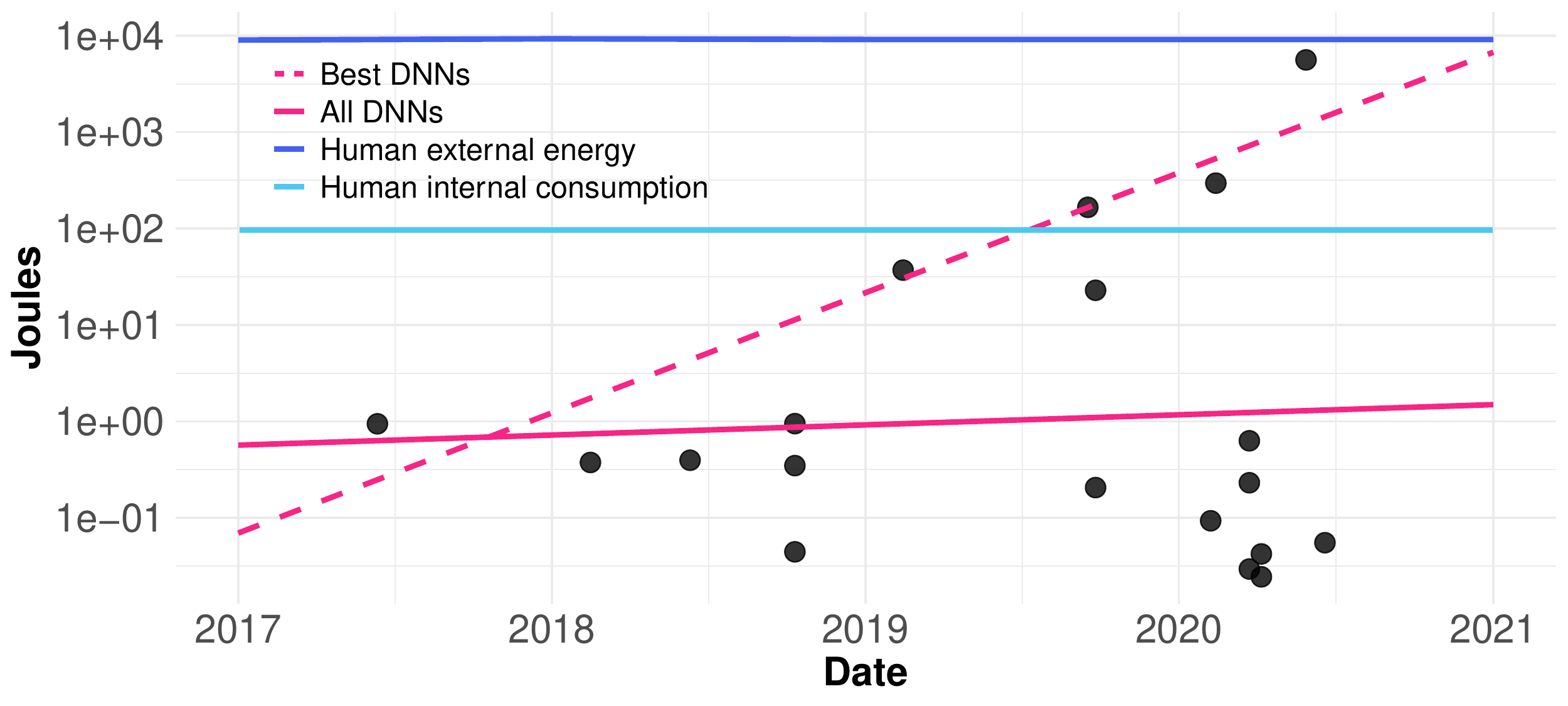}
\caption{\label{fig:joules_year_human_nlp}Estimated Joules per forward pass (e.g., one prediction) compared to human 
consumption in 1s (NLP).}
\end{figure}

In order to put this inference energy consumption in context we calculate the value of average human body energy consumption (we will refer to it as somatic or internal consumption) in one second and the average energy that a human being consumes in one second with all their commodities (we will refer to it as external consumption). The internal consumption is calculated assuming 
2000 
KCal 
per person day, and converting this to Joules/s, giving approximately 100 Joules/s. 
The external consumption is the sum of total energy consumption, including electricity, transport and heating, using the USA as a reference \citep{owidenergy}. 
This suggests 79,897 Kwh/year in 2019, 
which is approximately 10,000 Joules every second.
The comparison of these two references with the trends can be seen in Fig.~\ref{fig:joules_year_human} (CV). As we see, the energy consumed for one inference of the best models approaches the energy consumed by the human body in one second but stills far from the external energy consumed in one second. If each human did an AI-based decision implying a  forward pass every second during the whole day (and night), this would be still well below their internal consumption. However, AI-based decisions are becoming more ubiquitous. For instance, a self-driving car or a surveillance camera may be making many forward passes per second. For NLP, the trends are similar but the best models are growing much faster, as we see in Fig.~\ref{fig:joules_year_human_nlp}, while the regular models may even decrease. Here, the interpretation in terms of how many decisions are made in a second is also hard to determine. For instance, a language model interfaced by a human does not require more than the basic 128-token windows per second. However, many applications of language models can process data without interacting with humans at a much higher speed.

\section{Discussion and Future Work}\label{section:discussion}

In this work we have combined the analysis of several elements about AI, compute and energy consumption that allow us to have a different and more comprehensive perspective about the energy impact of AI. The most distinctive element of our analysis is that we focus on inference cost, which is usually lower than the training cost when both are reported in research papers, but because of multiplicative factors, it is much higher overall. 
Many DNN models are trained once and applied millions of times (forward passes).

Our findings are very different from the unbridled exponential growth that is usually reported when just looking at the number of parameters of new deep learning models \citep{hestness2017deep,kaplan2020scaling,henighan2020scaling}. 
When we focus on inference costs of these networks, the energy that is associated is not growing so fast, because of several factors that partially compensate the growth, such as algorithmic improvements, hardware specialisation and hardware consumption efficiency. The gap gets closer when we analyse those models that settle, i.e., those models whose implementation become very popular one or two years after the breakthrough algorithm was introduced. These {\em general-use} models can achieve systematic growth in performance at an almost constant energy consumption. 
The main conclusion is that even if the {\em energy} used by AI were kept constant, the improvement in performance could be sustained with algorithmic improvements and fast increase in the number of parameters. 

This conclusion has an important limitation. It assumes a constant multiplicative factor. As more and more devices use AI (locally or remotely) the energy consumption can escalate just by means of increased penetration, in the same way that cars have become more efficient in the past two decades but there are many more cars in the world today.

We hope this paper contributes to the increasing debate about AI and energy consumption by analysing the inference costs. As these are dominated by multiplicative factors, this should encourage not only AI researchers but economists and social scientists to participate in this analysis.  
Future studies would be enriched by socio-economic indicators about the use of AI (the degree of penetration), the cost of energy and devices as well as the carbon footprint per Joule \citep{europa-footprint}. Similarly, comparing energy consumption by AI and trends in  human salaries could help determine where automation \citep{tolan2021measuring} becomes cost effective in economic terms. 

Finally, this paper has many limitations that originate from the limited information reported in scientific papers. Many papers include the number of hyperparameters, but it is less common to have complete information about FLOPs and energy consumption. It is even rarer when looking for inference costs. This information is not only necessary for the transparency of the field but it is of utmost relevance for producing studies such as the one we have presented here, with a larger number of benchmarks and models. Also, it is important that new techniques are reported with new but also old benchmarks, so that we can have larger temporal windows where we can analyse the evolution of the field. We hope that future studies can build on this one and better publishing practices.

\bibliographystyle{abbrvnat}
\bibliography{biblio}

\clearpage

\appendix

\section{Appendix}

In this technical appendix we include some supplementary material  giving detailed information about 1) differences between FLOPs and FLOPS; 2) methodological details for CV and NLP models used in our analyses; 3) benchmarks addresed; 4) hardware specifics regarding precision; 5) further analysis for performance and compute in NLP tasks; 6) FLOPS estimation procedures; 7) Results for the GLUE benchmark; and 8) GPU consumption data. 

\subsection{FLOPs vs FLOPS}

When dealing about computing effort and computing speed (hardware performance), terminology is usually confusing. The term `compute' is usually ambiguous, sometimes applied for a number of operations or the number of operations per second. However, it is important to clarify what kind of operations and the acronyms for them. In this regard, we will use the acronym FLOPS to measure hardware performance, by referring to the number of floating point operations {\em per second}, as standardised in the industry, while FLOPs will be applied to the amount of computation for a given task (e.g., a prediction or inference pass), by referring to the number of operations, counting a multiply-add operation pair as two operations. 

For instance, we found out that the acronym FLOP may be misleading. By FLOP, we mean one floating point operation, a measure of the amount of compute (computing effort) and by FLOPS, we mean  floating point operations {\em per second}, i.e., FLOPS = FLOP/s. However, many papers, especially CV papers, use the terms FLOPs and FLOPS to refer to the number of operations, but we will be just use FLOPs as the plural of FLOP, never as FLOPS.  Then there is the question of what a FLOP is. When dealing with DNN, this is usually associated with the number of multiply-add operations, even there are other type of operations involved when executing a DNN. This is done this way because it is usually a good 
estimation \citep{how_fast,clark2020electra}. More specifically, we will count one fused multiply-add operation as 2 FLOPs (note the lowercase `s'). Hardware manufacturers count them in this manner \citep{nvidia_flops}, because in fact there are two mathematical operations. However, CV research papers count a multiply-add operation as only one operation. In this case, we will multiply the number of operations reported by 2. In sum, the acronym FLOPS will be applied to measure hardware performance, by referring to the number of floating point operations {\em per second}, as standardised in the industry, while FLOPs will be applied to the amount of computation for a given task (e.g., a prediction or inference pass), by referring to the number of operations, counting a multiply-add operation pair as two operations.

\subsection{Methodology Details for CV Models}

Accuracy and FLOPs metrics were collected carefully, taking into account that there are different sampling techniques to reach a given accuracy. For instance, in the AlexNet paper \citep{krizhevsky2012ImageNet}, to classify a single image they make 10 predictions, they take 10 different crops\footnote{Cropping is a common image manipulation process: while cropping the middle square (down-sampling) from input images is a good practice for data preparation, random cropping is also a good practice for train-data augmentation}  from the original image and average the 10 predictions to get the final prediction. While this is a useful trick, it is not fair to compare the accuracy of a model achieved with 10 crops with another achieved with 1 crop. Furthermore, the use of several crops or other kinds of repetitions is problematic, as the papers usually report the number of FLOPs for one forward pass\footnote{A "forward pass" refers to calculation process, values of the output layers from the inputs data. It's traversing through all neurons from first to last layer. A loss function is calculated from the output values.} (if 10 forward passes are needed to make a single prediction, then the FLOPs should be multiplied by 10). For these reasons we only report 1-crop accuracy for all models, to make a meaningful comparison.

Note that the FLOPs also depend of the input image resolution: the higher the image resolution, the more operations (FLOPs) are required to process it. Some researchers report results with different image resolutions \citep{simonyan2015deep,zhai2021scaling}, and sometimes it is not clear which resolution the results are reported for. In these cases, we need to investigate until we find that information. 
In sum, all the collected FLOPs in this work are for a forward pass with the resolution used for inference. The selected models and their values are shown in Table \ref{tab:modelDetails}. 

\renewcommand{\arraystretch}{0.99}
\setlength{\tabcolsep}{2pt}
\begin{table*}[!h]
\centering
\resizebox{\textwidth}{!}{%
\begin{tabular}{lcccccc}
\toprule
\textbf{Model} & \textbf{Top-1 Acc.} & \textbf{Params (M)} & \textbf{GFLOPs} & \textbf{Extra Data} & \textbf{Date} & \textbf{Architecture} \\ 
\midrule
AlexNet \citep{krizhevsky2012ImageNet} & 56.52 \citep{pytroch} & 61.00 $\dagger$ & 1.42 $\dagger$ & No & 01/06/2012 & CNN \\
ZFNet-b \citep{zeiler2013visualizing} & 63.63 \citep{cvpytroch} & 107.63 \citep{cvpytroch} & 4.96 \citep{cvpytroch} & No & 11/11/2013 & CNN \\
ZFNet \citep{zeiler2013visualizing} & 60.21 \citep{cvpytroch} & 62.36 \citep{cvpytroch} & 2.34 \citep{cvpytroch} & No & 12/11/2013 & CNN \\
VGG-19 \citep{simonyan2015deep} & 72.37 \citep{pytroch} & 144.00 & 39.34 $\dagger$ & No & 04/09/2014 & CNN \\
VGG-16 \citep{simonyan2015deep} & 71.59 \citep{pytroch} & 138.00 & 31.00 $\dagger$ & No & 04/09/2014 & CNN \\
Inception V1/GoogleLeNet \citep{szegedy2014going} & 69.77 \citep{pytroch} & 6.80 & 3.00 & No & 17/09/2014 & CNN \\
Inception V2/Incepton BN \citep{ioffe2015batch} & 74.80 & 11.29 \citep{cvpytroch} & 4.10 \citep{cvpytroch} & No & 11/02/2015 & CNN \\
Inception V3 \citep{szegedy2015rethinking} & 78.80 & 23.83 & 11.48 & No & 02/12/2015 & CNN \\
ResNet-50 \citep{he2015deep} & 75.30 \citep{drnGitHub} & 26.00 \citep{kerasApps} & 7.60 & No & 10/12/2015 & CNN \\
ResNet-101 \citep{he2015deep}& 76.40 \citep{drnGitHub} & 45.00 \citep{kerasApps} & 15.20 & No & 10/12/2015 & CNN \\
ResNet-152 \citep{he2015deep}& 77.00 \citep{drnGitHub} & 60.00 \citep{kerasApps} & 22.60 & No & 10/12/2015 & CNN \\
Inception V4 \citep{szegedy2016inceptionv4} & 80.00 & 42.68 \citep{cvpytroch} & 24.60 \citep{cvpytroch} & No & 23/02/2016 & CNN \\
Inception ResNet V2 \citep{szegedy2016inceptionv4} & 80.10 & 55.84 \citep{cvpytroch} & 26.38 \citep{cvpytroch} & No & 23/02/2016 & CNN \\
Densenet-121 \citep{huang2018densely} & 74.98 & 7.98 \citep{cvpytroch} & 5.74 \citep{cvpytroch} & No & 25/08/2016 & CNN \\
Densenet-169 \citep{huang2018densely} & 76.20 & 14.15 \citep{cvpytroch} & 6.80 \citep{cvpytroch} & No & 25/08/2016 & CNN \\
Densenet-201 \citep{huang2018densely} & 77.42 & 20.01 \citep{cvpytroch} & 8.68 \citep{cvpytroch} & No & 25/08/2016 & CNN \\
Xception \citep{chollet2017xception} & 79.00 & 22.86 & 16.80 \citep{cvpytroch} & No & 07/10/2016 & CNN \\
ResNeXt-50 (32x4d) \citep{xie2017aggregated} & 77.80 & 25.00 & 8.40 & No & 16/11/2016 & CNN \\
ResNeXt-101 (64x4d) \citep{xie2017aggregated} & 79.60 & 83.46 & 31.20 $\dagger$ & No & 16/11/2016 & CNN \\
MobileNet \citep{howard2017mobilenets} & 70.60 & 4.20 & 1.14 & No & 17/04/2017 & CNN \\
ShuffleNet x1.0 (g=8) \citep{zhang2017shufflenet} & 67.60 & 2.43 \citep{cvpytroch} & 0.28 & No & 04/07/2017 & CNN \\
DPN-131 (40 × 4d) \citep{chen2017dual} & 80.07 & 79.50 & 32.00 & No & 06/07/2017 & CNN \\
DPN-98 (40 × 4d) \citep{chen2017dual} & 79.80 & 61.70 & 23.40 & No & 06/07/2017 & CNN \\
DPN-92 (32 × 3d) \citep{chen2017dual} & 79.30 & 37.80 & 13.00 & No & 06/07/2017 & CNN \\
NASNet-A (6 @ 4032) \citep{zoph2018learning} & 82.70 & 88.90 & 47.60 & No & 21/07/2017 & CNN \\
NASNet-A (7 @ 1920) \citep{zoph2018learning} & 80.80 & 22.60 & 9.86 & No & 21/07/2017 & CNN \\
SENet-154 \citep{hu2019squeezeandexcitation} & 81.32 & 115.09 \citep{cvpytroch} & 41.50 \citep{cvpytroch} & No & 05/09/2017 & CNN \\
PNASNet-5 (N = 4, F = 216) \citep{liu2018progressive} & 82.90 & 86.10 & 50.00 & No & 02/12/2017 & CNN \\
PNASNet-5 (N = 3, F = 54) \citep{hu2019squeezeandexcitation} & 74.20 & 5.10 & 1.18 & No & 02/12/2017 & CNN \\
MobileNetV2 \citep{sandler2019mobilenetv2} & 72.00 & 3.40 & 0.60 & No & 13/01/2018 & CNN \\
MobileNetV2 1.4 \citep{sandler2019mobilenetv2} & 74.70 & 6.90 & 1.18 & No & 13/01/2018 & CNN \\
AmoebaNet-A (N=6, F=190) \citep{real2019regularized} & 82.80 & 86.70 & 46.20 & No & 05/02/2018 & CNN \\
AmoebaNet-A (N=6, F=448) \citep{real2019regularized} & 83.90 & 469.00 & 208.00 & No & 05/02/2018 & CNN \\
ResNeXt-101 32×32d \citep{mahajan2018exploring} & 85.10 & 466.00 & 174.00 & Instagram 940M & 02/05/2018 & CNN \\
ResNeXt-101 32×48d \citep{mahajan2018exploring} & 85.40 & 829.00 & 306.00 & Instagram 940M & 02/05/2018 & CNN \\
ShuffleNetV2 x1.0 \citep{ma2018shufflenet} & 69.40 & 2.28 \citep{cvpytroch} & 0.30 & No & 30/07/2018 & CNN \\
ResNeXt-101 32x16d \citep{yalniz2019billionscale,billionscaleGitHub} & 84.80 & 193.00 & 72.00 & Custom 940M & 02/05/2019 & CNN \\
ResNeXt-101 32x8d \citep{yalniz2019billionscale,billionscaleGitHub}& 84.30 & 88.00 & 32.00 & Custom 940M & 02/05/2019 & CNN \\
ResNeXt-50 32x4d \citep{yalniz2019billionscale,billionscaleGitHub} & 82.20 & 25.00 & 8.00 & Custom 940M & 02/05/2019 & CNN \\
EfficientNet-B0 \citep{tan2020efficientnet} & 77.10 & 5.30 & 0.78 & No & 28/05/2019 & CNN \\
EfficientNet-B1 \citep{tan2020efficientnet} & 79.10 & 7.80 & 1.40 & No & 28/05/2019 & CNN \\
EfficientNet-B2 \citep{tan2020efficientnet} & 80.10 & 9.20 & 2.00 & No & 28/05/2019 & CNN \\
EfficientNet-B3 \citep{tan2020efficientnet} & 81.60 & 12.00 & 3.60 & No & 28/05/2019 & CNN \\
EfficientNet-B4 \citep{tan2020efficientnet} & 82.90 & 19.00 & 8.40 & No & 28/05/2019 & CNN \\
EfficientNet-B5 \citep{tan2020efficientnet} & 83.60 & 30.00 & 19.80 & No & 28/05/2019 & CNN \\
EfficientNet-B6 \citep{tan2020efficientnet} & 84.00 & 43.00 & 38.00 & No & 28/05/2019 & CNN \\
EfficientNet-B7 \citep{tan2020efficientnet} & 84.30 & 66.00 & 74.00 & No & 28/05/2019 & CNN \\
NoisyStudent-B0 \citep{xie2020selftraining} & 78.80 & 5.30 & 0.78 & JFT 300M & 11/11/2019 & CNN \\
NoisyStudent-B1 \citep{xie2020selftraining} & 81.50 & 7.80 & 1.40 & JFT 300M & 11/11/2019 & CNN \\
NoisyStudent-B2 \citep{xie2020selftraining} & 82.40 & 9.20 & 2.00 & JFT 300M & 11/11/2019 & CNN \\
NoisyStudent-B3 \citep{xie2020selftraining} & 84.10 & 12.00 & 3.60 & JFT 300M & 11/11/2019 & CNN \\
NoisyStudent-B4 \citep{xie2020selftraining} & 85.30 & 19.00 & 8.40 & JFT 300M & 11/11/2019 & CNN \\
NoisyStudent-B5 \citep{xie2020selftraining} & 86.10 & 30.00 & 19.80 & JFT 300M & 11/11/2019 & CNN \\
NoisyStudent-B6 \citep{xie2020selftraining} & 86.40 & 43.00 & 38.00 & JFT 300M & 11/11/2019 & CNN \\
NoisyStudent-B7 \citep{xie2020selftraining} & 86.90 & 66.00 & 74.00 & JFT 300M & 11/11/2019 & CNN \\
NoisyStudent-L2 \citep{xie2020selftraining} & 88.40 & 480.00 & 1040.00 $\ast$ & JFT 300M & 11/11/2019 & CNN \\
FixEfficientNet-L2 \citep{touvron2020fixing} & 88.50 & 480.00 & 585.00 $\ast$ & JFT 300M & 18/03/2020 & CNN \\
FixEfficientNet-B7 \citep{touvron2020fixing} & 85.30 & 66.00 & 82.00 $\ast$ & No & 18/03/2020 & CNN \\
FixEfficientNet-B0 \citep{touvron2020fixing} & 79.30 & 5.30 & 1.60 $\ast$ & No & 18/03/2020 & CNN \\
Meta Pseudo Labels L2 \citep{pham2021meta} & 90.20 & 480.00 & 1040.00 $\ast$ & JFT 300M & 23/03/2020 & CNN \\
ResNeSt-269 \citep{zhang2020resnest} & 84.50 & 111.00 & 155.8 $\dagger$ & No & 19/04/2020 & CNN \\
ResNeSt-200 \citep{zhang2020resnest} & 83.90 & 70.00 & 71.56 $\dagger$ & No & 19/04/2020 & CNN \\
ResNeSt-50 \citep{zhang2020resnest} & 81.13 & 27.50 & 10.78 & No & 19/04/2020 & CNN \\
ViT-L/16 \citep{dosovitskiy2021image} & 85.30 & 304.00 \citep{tan2021efficientnetv2} & 384.00 \citep{tan2021efficientnetv2} & ImageNet 21k & 22/10/2020 & Transformer \\
ViT-L/16 \citep{dosovitskiy2021image} & 87.12 & 304.00 \citep{tan2021efficientnetv2} & 384.00 \citep{tan2021efficientnetv2} & JFT 300M & 22/10/2020 & Transformer \\
ViT-B/16 \citep{dosovitskiy2021image} & 84.60 \citep{tan2021efficientnetv2} & 87.00 \citep{tan2021efficientnetv2} & 112.00 \citep{tan2021efficientnetv2} & ImageNet 21k & 22/10/2020 & Transformer \\
DeiT-small \citep{touvron2021training, DeiTGitHub} & 79.90 & 22.00 & 9.20 \citep{yuan2021tokenstotoken} & No & 23/12/2020 & Transformer \\
DeiT-small-Distilled \citep{touvron2021training, DeiTGitHub} & 81.20 & 22.00 & 9.40 \citep{yuan2021tokenstotoken} & No & 23/12/2020 & Transformer \\
DeiT-base \citep{touvron2021training, DeiTGitHub} & 81.80 & 86.00 & 36.00 \citep{tan2021efficientnetv2} & No & 23/12/2020 & Transformer \\
DeiT-base-384 \citep{touvron2021training, DeiTGitHub} & 82.90 & 86.00 & 112.00 \citep{tan2021efficientnetv2} & No & 23/12/2020 & Transformer \\
BotNet-T7 \citep{srinivas2021bottleneck} & 84.70 & 75.00 & 92.00 & No & 27/01/2021 & Hybrid \\
BotNet-T5 \citep{srinivas2021bottleneck} & 83.50 & 75.10 & 38.60 & No & 27/01/2021 & Hybrid \\
T2T-ViTt-14 \citep{yuan2021tokenstotoken} & 81.70 & 21.50 & 12.20 & No & 28/01/2021 & Transformer \\
T2T-ViTt-19 \citep{yuan2021tokenstotoken} & 82.20 & 39.20 & 19.60 & No & 28/01/2021 & Transformer \\
T2T-ViTt-24 \citep{yuan2021tokenstotoken} & 82.60 & 64.10 & 30.00 & No & 28/01/2021 & Transformer \\
NFNet-F4+ \citep{brock2021highperformance} & 89.20 & 527.00 & 734.00 & JFT 300M & 11/02/2021 & CNN \\
NFNet-F0 \citep{brock2021highperformance} & 83.60 & 71.50 & 24.76 & No & 11/02/2021 & CNN \\
NFNet-F6+SAM \citep{brock2021highperformance} & 86.50 & 438.40 & 754.56 & No & 11/02/2021 & CNN \\
Swin-B 224 \citep{liu2021swin} & 85.20 & 88.00 & 30.80 & ImageNet 21k & 25/03/2021 & Transformer \\
Swin-B 384 \citep{liu2021swin} & 86.00 & 88.00 & 94.00 & ImageNet 21k & 25/03/2021 & Transformer \\
Swin-L \citep{liu2021swin} & 86.40 & 197.00 & 207.80 & ImageNet 21k & 25/03/2021 & Transformer \\
CrossViT-15 \citep{chen2021crossvit} & 81.50 & 27.40 & 11.60 & No & 27/03/2021 & Transformer \\
CrossViT-18 \citep{chen2021crossvit} & 82.50 & 43.30 & 18.06 & No & 27/03/2021 & Transformer \\
CaiT-S36 \citep{touvron2021going} & 83.30 & 68.00 & 27.80 & No & 31/03/2021 & Transformer \\
CaiT-S36 dist \citep{touvron2021going} & 84.00 & 68.00 & 27.80 & No & 31/03/2021 & Transformer \\
CaiT-S24-384 dist \citep{touvron2021going} & 85.10 & 46.90 & 64.40 & No & 31/03/2021 & Transformer \\
CaiT-M48-448 dist \citep{touvron2021going} & 86.50 & 356.00 & 659.20 & No & 31/03/2021 & Transformer \\
EfficientNetV2-S \citep{tan2021efficientnetv2} & 83.90 & 24.00 & 17.60 & No & 01/04/2021 & CNN \\
EfficientNetV2-M \citep{tan2021efficientnetv2} & 85.10 & 55.00 & 48.00 & No & 01/04/2021 & CNN \\
EfficientNetV2-L \citep{tan2021efficientnetv2} & 85.70 & 121.00 & 106.00 & No & 01/04/2021 & CNN \\
EfficientNetV2-S \citep{tan2021efficientnetv2} & 85.00 & 24.00 & 17.60 & ImageNet 21k & 01/04/2021 & CNN \\
EfficientNetV2-M \citep{tan2021efficientnetv2} & 86.10 & 55.00 & 48.00 & ImageNet 21k & 01/04/2021 & CNN \\
EfficientNetV2-L \citep{tan2021efficientnetv2} & 86.80 & 121.00 & 106.00 & ImageNet 21k & 01/04/2021 & CNN \\
ViT-G/14 \citep{zhai2021scaling} & 90.45 & 1843.00 & 5270.00 $\ast$ & JFT 3B & 08/06/2021 & Transformer \\ \bottomrule
\end{tabular}%
}
\caption{CV models data set. A citation next to a given value means that this value is extracted from that source, otherwise the values are from the paper (cited in model column). The symbol $\dagger$ means that this value was obtained or checked from a model implementation using model analysis tools, and the symbol $\ast$ means that we estimated the value.\label{tab:modelDetails}}
\end{table*}

\subsection{Methodology Details for NLP Models}

As previously stated, for NLP models we just included all the models since 2017 for which we find inference compute estimation. Many papers do not explain how they count FLOPs (as single mathematical operations or single hardware instructions), but we ultimately found out this information  explained in \citep{clark2020electra}. We compare the presented numbers with estimations in other publications (we compare the numbers for repeated and similar models) and we see that these numbers are very similar. We assume that the other authors follow this as the standard procedure to count FLOPs. In NLP, they count FLOPs as single mathematical operations and not as a single hardware instructions (like in CV). The important thing is that we use the same approach in all the NLP models, as the comparison and analysis will be intra-domain and never inter-domain.

\subsection{Datasets}

\subsubsection{ImageNet} 
ImageNet is the most used dataset in the last decade for training and evaluating CV models. The full dataset consists of 14,197,122 images distributed in 21,841 classes. Researchers refer to this dataset as ImageNet21k or ImageNet22k. However, researchers commonly use a subset of the full ImageNet dataset. This subset consists of 1.2 million images for training and 50,000 images for validation distributed in 1,000 classes. This subset was released for ImageNet Large Scale Visual Recognition Challenge 2012 (ILSVRC2012) and is usually referred as ImageNet1k or just as ImageNet. In 2012 the AlexNet model \citep{krizhevsky2012ImageNet} won the ILSVRC 2012 Image Classification with an impressive result, outperforming the other models by large margin. AlexNet was the first DNN to win this competition. Since then many other DNNs have been created for image classification. 

\subsubsection{GLUE}
The General Language Understanding Evaluation (GLUE) benchmark \citep{wang2019glue} is a collection of resources for evaluating and analysing the performance of models across a diverse range of existing NLP tasks with the goal of driving ``research in the development of general and robust natural language understanding systems". The collection in GLUE  consists of nine ``difficult and diverse" tasks, mostly adopted from existing datasets. The tasks involve sentiment analysis, acceptability, paraphrasing, natural language inference and coreference resolution.  GLUE is model-agnostic, but it incentivises sharing knowledge across tasks (using parameter sharing or other transfer learning techniques) due to the  limited training data for certain tasks.

\subsection{Hardware data compilation: floating point precision details}

At the end of 2017 Nvidia launched GPUs with new features for AI acceleration (improved lower precision performance and tensor cores, which can improve low-precision calculations) \citep{volta_white_paper}. For instance, many new GPUs have accelerated FP16 operations through tensor cores (DNN can operate at low precision in many calculations without problems) and combine them with FP32 precision operations when is necessary. In this way we benefit from higher performance, maintaining calculation's precision. Nvidia specifies different FLOPS for FP16 and for tensor cores. Nowadays, frameworks as PyTorch and TensorFlow allow to train and infer with a DNN with mixed precision, i.e., taking advantage of the tensor cores, easily without practically any significant reduction in accuracy. Because of all this, we consider necessary to include the performance achieved with tensor cores in our analysis. 

Theoretical FLOPS using tensor cores are very high, but this increase in FLOPS does not correspond with the gain seen in practice for deep learning applications (maybe gaming is different). This is because it is not possible to use tensor cores for all operations. To solve the discrepancy between tensor core FLOPS and the real utilisation of these FLOPS, we calculate the speed up achieved for DNN when inference is done with mixed precision. We have looked for experimental results to adjust the tensor FP16/FP32 FLOPS to real performance improvement, the inference experimental results that we use are available in Nvidia NGC Catalog\footnote{\url{https://ngc.nvidia.com/catalog/resources}}. The collected data can be found in Table \ref{tab:gpu_speed_up_models}.

\renewcommand{\arraystretch}{0.99}
\setlength{\tabcolsep}{7pt}
\begin{table*}[!h]
\centering
\resizebox{0.8\textwidth}{!}{
\begin{tabular}{@{}clcccccc@{}}

\toprule
\textbf{Task} & \multicolumn{1}{l}{\textbf{Model}} & \textbf{Framework} & \textbf{Batch size} & \textbf{GPU} & \textbf{Presicion} & \textbf{Throughput} & \textbf{Speed-up} \\ \midrule
\multirow{36}{*}{\textbf{CV}}
 & efficientnet-b0 & PyTorch & 256 & V100 16GB & FP32 & 2968 & 1.00 \\
 & efficientnet-b0 & PyTorch & 256 & V100 16GB & Mixed & 6176 & 2.08 \\
 & efficientnet-b0 & PyTorch & 256 & A100 80GB & TF32 & 5154 & 1.74 \\
 & efficientnet-b0 & PyTorch & 256 & A100 80GB & Mixed & 10239 & 3.45 \\
 & efficientnet-b4 & PyTorch & 128 & V100 16GB & FP32 & 376 & 1.00 \\
 & efficientnet-b4 & PyTorch & 128 & V100 16GB & Mixed & 843 & 2.24 \\
 & efficientnet-b4 & PyTorch & 128 & A100 80GB & TF32 & 700 & 1.86 \\
 & efficientnet-b4 & PyTorch & 128 & A100 80GB & Mixed & 1418 & 3.77 \\
 & ResNeXt101-32x4d & PyTorch & 256 & V100 16GB & FP32 & 533 & 1.00 \\
 & ResNeXt101-32x4d & PyTorch & 256 & V100 16GB & Mixed & 1746 & 3.28 \\
 & ResNeXt101-32x4d & PyTorch & 256 & T4 16GB & FP32 & 161 & 1.00 \\
 & ResNeXt101-32x4d & PyTorch & 256 & T4 16GB & Mixed & 598 & 3.71 \\
 & ResNet v1.5 & PyTorch & 256 & V100 16GB & FP32 & 1261 & 1.00 \\
 & ResNet v1.5 & PyTorch & 256 & V100 16GB & Mixed & 3382 & 2.68 \\
 & ResNet v1.5 & PyTorch & 256 & T4 16GB & FP32 & 415 & 1.00 \\
 & ResNet v1.5 & PyTorch & 256 & T4 16GB & Mixed & 1198 & 2.89 \\
 & ResNet v1.5 & TensorFlow & 256 & V100 16GB & FP32 & 1348.52 & 1.00 \\
 & ResNet v1.5 & TensorFlow & 256 & V100 16GB & Mixed & 2742.14 & 2.03 \\
 & ResNet v1.5 & TensorFlow & 256 & A100 40GB & TF32 & 1911.96 & 1.42 \\
 & ResNet v1.5 & TensorFlow & 256 & A100 40GB & Mixed & 3229.32 & 2.39 \\
 & ResNet v1.5 & TensorFlow & 256 & T4 16GB & FP32 & 425.72 & 1.00 \\
 & ResNet v1.5 & TensorFlow & 256 & T4 16GB & Mixed & 993.39 & 2.33 \\
 & SSD v1.1 & PyTorch & 32 & V100 16GB & FP32 & 271.73 & 1.00 \\
 & SSD v1.1 & PyTorch & 32 & V100 16GB & Mixed & 438.85 & 1.62 \\
 & SSD v1.1 & PyTorch & 32 & A100 40GB & TF32 & 548.75 & 2.02 \\
 & SSD v1.1 & PyTorch & 32 & A100 40GB & Mixed & 910.17 & 3.35 \\
 & UNet Industrial & TensorFlow & 16 & V100 16GB & FP32 & 250.23 & 1.00 \\
 & UNet Industrial & TensorFlow & 16 & V100 16GB & Mixed & 469.27 & 1.88 \\
 & UNet Industrial & TensorFlow & 16 & A100 40GB & TF32 & 424.57 & 1.70 \\
 & UNet Industrial & TensorFlow & 16 & A100 40GB & Mixed & 823.46 & 3.29 \\
 & SE-ResNeXt101-32x4d & TensorFlow & 128 & V100 16GB & FP32 & 460.82 & 1.00 \\
 & SE-ResNeXt101-32x4d & TensorFlow & 128 & V100 16GB & Mixed & 1102 & 2.39 \\
 & SE-ResNeXt101-32x4d & TensorFlow & 128 & A100 40GB & TF32 & 802.64 & 1.74 \\
 & SE-ResNeXt101-32x4d & TensorFlow & 128 & A100 40GB & Mixed & 1728.27 & 3.75 \\
 & SE-ResNeXt101-32x4d & TensorFlow & 128 & T4 16GB & FP32 & 105.16 & 1.00 \\
 & SE-ResNeXt101-32x4d & TensorFlow & 128 & T4 16GB & Mixed & 195.17 & 1.86 \\ \midrule
\multirow{22}{*}{\textbf{NLP}}
 & BERT-LARGE & TensorFlow & 8 & V100 16GB & FP32 & 44.03 & 1.00 \\
 & BERT-LARGE & TensorFlow & 8 & V100 16GB & Mixed & 168.34 & 3.82 \\
 & BERT-LARGE & TensorFlow & 8 & A100 80GB & TF32 & 241.68 & 5.49 \\
 & BERT-LARGE & TensorFlow & 8 & A100 80GB & Mixed & 342.22 & 7.77 \\
 & BERT-LARGE & TensorFlow & 8 & T4 16GB & FP32 & 16.04 & 1.00 \\
 & BERT-LARGE & TensorFlow & 8 & T4 16GB & Mixed & 62.99 & 3.93 \\
 & BERT-Base & TensorFlow & 8 & V100 16GB & FP32 & 146.15 & 1.00 \\
 & BERT-Base & TensorFlow & 8 & V100 16GB & Mixed & 504.24 & 3.45 \\
 & BERT-Base & TensorFlow & 8 & A100 80GB & TF32 & 645.88 & 4.42 \\
 & BERT-Base & TensorFlow & 8 & A100 80GB & Mixed & 846.81 & 5.79 \\
 & BERT-Base & TensorFlow & 8 & T4 16GB & FP32 & 51.33 & 1.00 \\
 & BERT-Base & TensorFlow & 8 & T4 16GB & Mixed & 192.61 & 3.75 \\
 & Transformer-XL & TensorFlow & 32 & V100 16GB & FP32 & 8555.6 & 1.00 \\
 & Transformer-XL & TensorFlow & 32 & V100 16GB & Mixed & 11215.5 & 1.31 \\
 & Transformer-XL & TensorFlow & 32 & A100 40GB & TF32 & 19434.5 & 2.27 \\
 & Transformer-XL & TensorFlow & 32 & A100 40GB & Mixed & 21854.7 & 2.55 \\
 & Transformer-XL & TensorFlow & 32 & T4 16GB & FP32 & 3439.1 & 1.00 \\
 & Transformer-XL & TensorFlow & 32 & T4 16GB & Mixed & 6174.3 & 1.80 \\
 & Transformer & PyTorch & 10240 & V100 16GB & FP32 & 3782 & 1.00 \\
 & Transformer & PyTorch & 10240 & V100 16GB & Mixed & 7464 & 1.97 \\
 & Transformer & PyTorch & 10240 & A100 40GB & TF32 & 7755 & 2.05 \\
 & Transformer & PyTorch & 10240 & A100 40GB & Mixed & 9653 & 2.55  \\ \bottomrule
\end{tabular}%
}
\caption{Throughput measures for V100, A100 and T4 GPUs on different Models.  The `speed-up' column is  the  speed-up  achieved  with  respect  to  FP32  throughput  using  different  precision  formats. A100 speed-up is calculated with respect to V100 FP32 throughput. The data is obtained from NVIDIA NGC catalog (https://ngc.nvidia.com/catalog/resources).}
\label{tab:gpu_speed_up_models}
\end{table*}

We do not include estimated mixed precision performance for all GPUs that support it because we have not found sufficient benchmarks for all GPUs to carry out an estimation. Also, we do not consider INT8 precision format because in many cases using this format leads to performance downgrade, and therefore the accuracy metric of the models should be adapted for a fair analysis. We perform a different estimation for CV and for NLP networks because these two kinds of networks operate in different ways and take different advantage of mixed precision. 
During training the speed-up from mixed precision in comparison to FP32 is usually of 2x for image models, and up to 4x for language models \citep{lamdaV100}. This is corroborated in information about some benchmarks on Nvidia blogs too \citep{amp}. 

\subsection{Hardware mixed precision speed-ups }

As we have discussed, theoretical FLOPS for tensor cores are very high, as we can see in Fig.~\ref{fig:gpus} in the main text. However, the performance for inference using tensor cores is not so high. For this reason we propose an estimation for the Nvidia GPUS: V100, A100 and T4 for CV models and for NLP models. For these calculations we collected inference data from NVIDIA NGC. The estimations for A100 are in relation to V100 because there is no data about FP32 for A100 (because FP32 is substituted by TF32 \footnote{\url{https://blogs.nvidia.com/blog/2020/05/14/tensorfloat-32-precision-format/}}, which is a precision format in between of FP32 and FP16), so we estimated the speed-up to V100 FP32 FLOPS.

\begin{table*}[!h]
\centering
\resizebox{0.7\columnwidth}{!}{%
\begin{tabular}{@{}clcc@{}}
\toprule
\textbf{GPU} & \textbf{Precision speed up} & \textbf{CV models} & \textbf{NLP models} \\ \midrule
V100 & Mixed speed up ratio to V100 FP32 & 2.27 & 2.64 \\\midrule
\multirow{2}{*}{A100} & TF32 speed up ratio to V100 FP32 & 1.75 & 3.56 \\
 & Mixed speed up ratio to V100 FP32 & 3.33 & 4.67 \\\midrule
T4 & Mixed speed up ratio to T4 FP32 & 2.7 & 3.16 \\ \bottomrule
\end{tabular}%
}
\caption{Mixed precision speed ups from experimental results for inference.\label{table:table_speed_up}}
\end{table*}

\subsection{Performance and compute (NLP)}

\begin{figure}[!h]
\centering
\includegraphics[width=0.65\columnwidth]{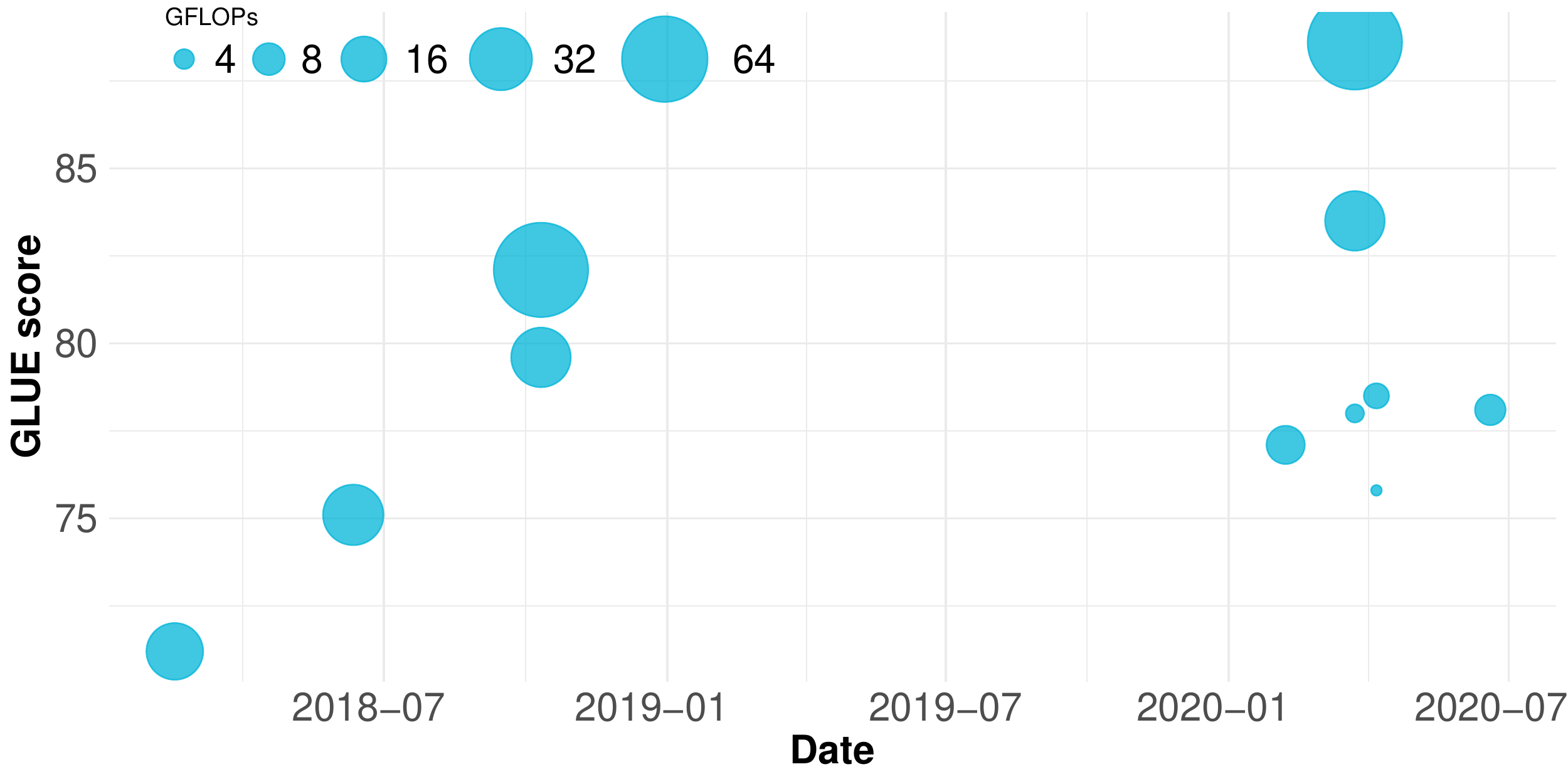}
\caption{\label{fig:date_glue}GFLOPs per token analysis for NLP models.}
\end{figure}

We represent the improvement on the GLUE score over the years as well as models inference GFLOPs (bubbles size) in Fig.~\ref{fig:date_glue}. GFLOPs are for single input of length 128, which is a reasonable sequence length for many use cases, being able to fit text messages or short emails. We can observe a very similar evolution to the evolution observed in ImageNet: SOTA models require a large number of FLOPs, but in a short period of time other models appear, which require much fewer FLOPs to reach the same score.

\subsection{FLOPS estimation for CV models}

\subsubsection{EfficientNet-Based Models FLOPs Estimation}
There are many EfficientNet variations, mostly using different input resolution or scaling. For these modifications, FLOPs are not always reported. In this work, we estimate them following the relation presented in Equation \ref{eq:relationflops}

\begin{equation}
\mbox{FLOPs} \propto d + w^2 + r^2
\label{eq:relationflops}
\end{equation}

\noindent for the following models:

\begin{itemize}
    \item \textbf{NoisyStudent-L2}: Having the scale factors of the networks (Table \ref{tab:scale_factors}) we estimate NoisyStudent-L2 FLOPs as shown in Equation \ref{eq:ns1}
    \begin{table}[!htbp]
    \centering
    \resizebox{0.5\columnwidth}{!}{%
    \begin{tabular}{lccc}
    \toprule
    \textbf{Model} & \textit{\textbf{w}} & \textit{\textbf{d}} & \textbf{Test Resolution} \\
    \midrule
    EfficientNet-B7 & 2 & 3.1 & $600 \times 600$ \\
    EfficientNet-L2 & 4.3 & 5.3 & $800 \times 800$\\ \bottomrule
    \end{tabular}%
    }
    \vspace{0.4cm}
    \caption{EfficientNet models architecture specifications obtained from \citep{xie2020selftraining}.}
    \label{tab:scale_factors}
    \end{table}
    \begin{equation}
    \begin{split}
    \mbox{NoisyStudent-L2 FLOPs} = & \\ 
    = \mbox{EfficientNet-B7 FLOPs} \cdot \mbox{$d_{\sigma} \cdot w_{\sigma}^2 \cdot r_{\sigma}^2$}
    \label{eq:ns1}
    \end{split}
    \end{equation}

    \noindent where $d_{\sigma}$, $w_{\sigma}$ and $r_{\sigma}$ are the scaled factors for, respectively, the network depth, width and input resolutions. By using the values from Table \ref{tab:scale_factors}, $d_{\sigma} = {5.3}/{3.1} = 1.7097$, $w_{\sigma} = {4.3}/{2} = 2.15$ and $r_{\sigma} = {800}/{600} = 1.3334$. Knowing that the GFLOPS for EfficientNet-B7 are 74, substituting in \ref{eq:ns1}, we obtain the estimation of $74 \mbox{ GFLOPs} \cdot 1.7097 \cdot 2.15^2 \cdot 1.3334^2 \approx 1040$ GFLOPS for NoisyStudent-L2.

    \item \textbf{Meta Pseudo Labels L2}: We use the estimation of NoisyStudent-L2 FLOPs for Meta Pseudo Labels L2, because it is the same model and only changes the training strategy.

    \item \textbf{FixEfficientNet-L2}: In FixEfficientNet-L2 they use a resolution of $600 \times 600$ for testing, so the estimation is the same as for NoisyStudent-L2 but without taking into account the resolution scaling ($r_{\sigma}$). Then, the estimated GFLOPS are $74 \mbox{ GFLOPs} \cdot 1.7097 \cdot 2.15^2 \approx 585$ GFLOPS.
    
    \item \textbf{FixEfficientNet-B7}: This model is the same as EfficientNet-B7 but using a slightly different resolution ($632 \times 632$). Therefore,  $r_{\sigma} = 632/600 = 1.0534$ and, thus we estimate $74 \mbox{ GFLOPs} \cdot 1.0534^2 \approx 82$ GFLOPs.
    
    \item \textbf{FixEfficientNet-B0}: This model is the same as EfficientNet-B0 but using a higher resolution ($320 \times 320$). Therefore,  $r_{\sigma} = 320/224 = 1.4286$ and, thus we estimate $0.78 \mbox{ GFLOPs} \cdot 1.4286^2 \approx 1.6$ GFLOPs.

\end{itemize}

\subsubsection{ViT-G/14 FLOPs Estimation}

In the paper \citep{zhai2021scaling} introducing the model, although authors provide the GFLOPs for $224 \times 224$ and $384 \times 384$ resolutions (see Table \ref{tab:vit_flops}), they also also use $518 \times 518$ resolution for ViT-G finetuning, so we assume they use the same resolution for testing too. ViT-G/14 is a vision transformer model, so the scale relation presented in \ref{eq:relationflops} do not apply for this kind of models. However, knowing the GFLOPs for $224 \times 224$ and $384 \times 384$, we may calculate how GFLOPs scale with resolution (given that $r_{\sigma}^2 = ({384}/{224})^2 = 2.9388$). In this regard, we calculate the GFLOPs ratio as ${2859.9}/{965.3} = 2.9627$ and we observe that GFLOPs scale quadratically with respect to resolution. Note, in this paper they report ``real" FLOPs and not multiply-add operations. Therefore, we recalculate $r_{\sigma} = {518}/{384} = 1.3490$ and multiply the GFLOPs for $384 \times 384$ resolution by this scale factor estimating $2859.9 \mbox{ GFLOPs} \cdot 1.3490^2 \approx 5270$ GFLOPs for the ViT-G/14 model.

 \begin{table}[!h]
\centering
\resizebox{0.35\columnwidth}{!}{%
\begin{tabular}{@{}ccc@{}}
\toprule
\multirow{2}{*}{\textbf{Model}} & \multicolumn{2}{c}{\textbf{GFLOPS}} \\ \cline{2-3} 
 & \textbf{$224 \times 224$} & \textbf{$384 \times 384$} \\ \midrule
ViT-G/14 & 965.3 & 2859.9 \\ \bottomrule
\end{tabular}%
}
\vspace{0.4cm}
\caption{ViT-G/14 GFLOPs from.}
\label{tab:vit_flops}
\end{table}

\subsection{NLP data}

Many times researchers report GLUE score without the punctuation on the WNLI task, because this task is problematic. We have marked which scores are reported without this task. Since there are 9 tasks in total, we consider that excluding one of them is not problematic for our analysis.

We did not find inference GFLOPs for the model Bert-Large, but we have ELECTRA-Large GFLOPs and this is actually the same model but following a different training strategy. In this sense, we use take ELECTRA-Large GFLOPs as BERT-Large GFLOPs. For ELMo we take GLUE ``dev-set" score because we do not found the score on the test set (we assume this score should be close to the test set). Values shown in Table \ref{tab:nlp_data}.

\begin{table*}[!h]
\centering
\resizebox{\textwidth}{!}{%
\begin{tabular}{lccccc}
\toprule
\textbf{Model} & \textbf{Input Tokens} & \textbf{GFLOPs} & \textbf{Params (M)} & \textbf{Date} & \textbf{GLUE test set} \\ \midrule
Transformer \citep{vaswani2017attention} & 512 & 54 \citep{git_nlp} & 65 & 12/06/2017 & - \\
ELMo \citep{peters2018deep} & 128 & 26 \citep{clark2020electra} & 96 & 15/02/2018 & 71.2 \citep{clark2020electra} $\clubsuit$ \\
GPT-1 \citep{radford2018improving} & 128 & 30 \citep{clark2020electra} & 117 & 11/06/2018 & 75.1 \citep{devlin2019bert} $\spadesuit$ \\
BERT Large \citep{devlin2019bert} & 128 & 79 & 335 $\ast$ & 11/10/2018 & 82.1 $\spadesuit$  \\
BERT-Small \citep{devlin2019bert} & 128 & 3.7 \citep{clark2020electra} & 14 & 11/10/2018 & -  \\
BERT-Base \citep{devlin2019bert} & 128 & 29 \citep{clark2020electra} & 110 & 11/10/2018 & 79.6 $\spadesuit$  \\
GPT-2 \citep{radford2019language} & 1024 & 3400 \citep{git_nlp} & 1500 & 14/02/2019 & - \\
Megatron \citep{shoeybi2020megatronlm} & 1024 & 18000 \citep{git_nlp} & 8300 & 17/09/2019 & -  \\
ALBERT-xxl \citep{lan2020albert} & 512 & 2500 \citep{git_nlp} & 235 & 26/09/2019 & -  \\
ALBERT-base \citep{lan2020albert} & 128 & 22.5 \citep{iandola2020squeezebert} & 12 & 26/09/2019 & - \\
Theseus 6/768 \citep{xu2020bertoftheseus} & 128 & 11.3 \citep{iandola2020squeezebert} & 66 & 07/02/2020 & 77.1 \citep{iandola2020squeezebert} \\
Microsoft T-NLG \citep{turing2020} & 1024 & 36000 \citep{git_nlp} & 17000 & 13/02/2020 & - \\
ELECTRA Large \citep{clark2020electra} & 128 & 79 \citep{git_nlp} & 335 & 23/03/2020 & 88.6 $\spadesuit$  \\
ELECTRA-Small \citep{clark2020electra} & 128 & 3.7 & 14 & 23/03/2020 & 78 $\spadesuit$  \\
ELECTRA-Base \citep{clark2020electra} & 128 & 29 & 110 & 23/03/2020 & 83.5 $\spadesuit$  \\
MobileBERT \citep{sun2020mobilebert} & 128 & 5.36 & 25.3 & 06/04/2020 & 78.5 $\spadesuit$ \\
MobileBERT tiny \citep{sun2020mobilebert} & 128 & 3.1 & 15.1 & 06/04/2020 & 75.8 $\spadesuit$ \\
GPT-3 \citep{brown2020language} & 2048 & 740000 \citep{git_nlp} & 175000 & 28/05/2020 & - \\
SqueezeBERT \citep{iandola2020squeezebert} & 128 & 7.42 & 51.1 & 19/06/2020 & 78.1 \\ \bottomrule
\end{tabular}%
}
\caption{NLP models data set. If there is a citation next to the GFLOPs value means that GFLOPs and Input Tokens values are extracted from that source, otherwise the values are from the paper (cited in the `Model' column). The symbol $\spadesuit$ means that GLUE score was calculated without punctuation on the WNLI task; the symbol $\ast$ means that we estimated the value and $\clubsuit$ means that GLUE score is for GLUE dev set instead of test set.}
\label{tab:nlp_data}
\end{table*}

\subsection{GPU consumption data}\label{sec:appendix_gpus_speed_up}

Tables \ref{tab:gpu_theorical} and \ref{tab:gpu_models} show further technical details regarding, respectively, the GPU's theoretical characteristics (compiled from the manufacturer's specification sheet and reference manuals), and their throughput and power consumption ``adapted", if necessary, to the specifics of CV or NLP tasks. 

\renewcommand{\arraystretch}{0.95}
\setlength{\tabcolsep}{1pt}
\begin{table}[!h]
\centering
\resizebox{0.85\columnwidth}{!}{%
\begin{tabular}{lcccccc}
\toprule
\textbf{GPU} & \textbf{Precision} & \textbf{TFLOPS} & \textbf{Watts} & \textbf{Launch date} & \textbf{Type} & \textbf{GFLOPS/Watt} \\ \midrule
GeForce  GTX 580 & FP32 & 1.58 & 244 & 09/11/2010 & Desktop & 6.48 \\
GeForce GTX 590 & FP32 & 2.49 & 365 & 24/03/2011 & Desktop & 6.82 \\
GeForce GTX 680 & FP32 & 3.09 & 195 & 22/03/2012 & Desktop & 15.85 \\
GeForce GTX 690 & FP32 & 5.62 & 300 & 29/04/2012 & Desktop & 18.73 \\
GeForce GTX 780 & FP32 & 4.16 & 250 & 23/04/2013 & Desktop & 16.62 \\
GeForce GTX 780 TI & FP32 & 5.35 & 250 & 07/11/2013 & Desktop & 21.38 \\
GeForce GTX Titan Black & FP32 & 5.65 & 250 & 18/02/2014 & Desktop & 22.58 \\
GeForce GTX Titan Z & FP32 & 8.12 & 375 & 28/05/2014 & Desktop & 21.66 \\
GeForce GTX 980 & FP32 & 4.98 & 165 & 18/09/2014 & Desktop & 30.19 \\
GeForce GTX 980 Ti & FP32 & 6.06 & 250 & 02/06/2015 & Desktop & 24.24 \\
GeForce GTX TITAN X & FP32 & 6.69 & 250 & 17/03/2015 & Desktop & 26.76 \\
GeForce GTX 1080 & FP32 & 8.87 & 180 & 26/05/2016 & Desktop & 49.29 \\
GeForce GTX 1080 Ti & FP32 & 11.34 & 250 & 10/03/2017 & Desktop & 45.36 \\
TITAN X Pascal & FP32 & 10.97 & 250 & 02/08/2016 & Desktop & 43.88 \\
TITAN XP & FP32 & 12.15 & 250 & 06/04/2017 & Desktop & 48.60 \\
GeForce RTX 2080 & FP32 & 10.07 & 215 & 20/09/2018 & Desktop & 46.84 \\
GeForce RTX 2080 Ti & FP32 & 13.45 & 250 & 20/09/2018 & Desktop & 53.80 \\
Nvidia Titan RTX & FP32 & 16.31 & 280 & 18/12/2018 & Desktop & 58.26 \\
GeForce RTX 3080 & FP32 & 29.80 & 320 & 01/09/2020 & Desktop & 93.13 \\
GeForce RTX 3090 & FP32 & 35.60 & 350 & 01/09/2020 & Desktop & 101.71 \\
GeForce RTX 2080 & FP16 & 20.14 & 215 & 20/09/2018 & Desktop & 93.67 \\
GeForce RTX 2080 Ti & FP16 & 26.90 & 250 & 20/09/2018 & Desktop & 107.60 \\
Nvidia Titan RTX & FP16 & 32.62 & 280 & 18/12/2018 & Desktop & 116.50 \\
GeForce RTX 3080 & FP16 & 29.80 & 320 & 01/09/2020 & Desktop & 93.13 \\
GeForce RTX 3090 & FP16 & 35.60 & 350 & 01/09/2020 & Desktop & 101.71 \\
GeForce RTX 2080 & FP16/FP32 Tensor & 40.30 & 215 & 20/09/2018 & Desktop & 187.44 \\
GeForce RTX 2080 Ti & FP16/FP32 Tensor & 56.90 & 250 & 20/09/2018 & Desktop & 227.60 \\
Nvidia Titan RTX & FP16/FP32 Tensor & 130.50 & 280 & 18/12/2018 & Desktop & 466.07 \\
GeForce RTX 3080 & FP16/FP32 Tensor & 59.50 & 320 & 01/09/2020 & Desktop & 185.94 \\
GeForce RTX 3090 & FP16/FP32 Tensor & 71.00 & 350 & 01/09/2020 & Desktop & 202.86 \\
Tesla K10 & FP32 & 4.58 & 225 & 01/05/2012 & Server & 20.36 \\
Tesla K20x & FP32 & 3.94 & 235 & 12/11/2012 & Server & 16.74 \\
Tesla K40 & FP32 & 5.04 & 235 & 08/10/2013 & Server & 21.45 \\
Tesla K80 & FP32 & 8.22 & 300 & 17/10/2014 & Server & 27.40 \\
Tesla M40 & FP32 & 6.84 & 250 & 10/10/2015 & Server & 27.36 \\
Tesla M60 & FP32 & 9.65 & 300 & 30/08/2015 & Server & 32.17 \\
Tesla P100 & FP16 & 21.20 & 300 & 20/05/2016 & Server & 70.67 \\
Tesla V100 & FP16 & 31.40 & 300 & 27/03/2018 & Server & 104.67 \\
A100 & FP16 & 78.00 & 400 & 14/04/2020 & Server & 195.00 \\
Tesla P100 & FP32 & 10.60 & 300 & 20/05/2016 & Server & 35.33 \\
Tesla V100 & FP32 & 15.70 & 300 & 27/03/2018 & Server & 52.33 \\
A100 & FP32 & 19.50 & 400 & 14/04/2020 & Server & 48.75 \\
A30 & FP32 & 10.30 & 165 & 12/04/2021 & Server & 62.42 \\
Tesla V100 & FP16/FP32 Tensor & 125.00 & 300 & 27/03/2018 & Server & 416.67 \\
A100 & FP16/FP32 Tensor & 312.00 & 400 & 14/04/2020 & Server & 780.00 \\
A30 & FP16/FP32 Tensor & 165.00 & 165 & 12/04/2021 & Server & 1000.00 \\
T4 & FP32 & 8.10 & 70 & 13/09/2018 & Server & 115.71 \\
T4 & FP16/FP32 Tensor & 65.00 & 70 & 13/09/2018 & Server & 928.57 \\ \bottomrule
\end{tabular}%
}
\caption{Nvidia GPUs theoretical data recopilation.}
\label{tab:gpu_theorical}
\end{table}

\renewcommand{\arraystretch}{0.99}
\begin{table}[!h]
\centering
\resizebox{0.85\columnwidth}{!}{%
\begin{tabular}{lccccccc}
\toprule
\textbf{Adapted} & \textbf{GPU} & \textbf{Precision} & \textbf{TFLOPS} & \textbf{Watts} & \textbf{Launch date} & \textbf{Type} & \textbf{GFLOPS/Watt} \\ \midrule
\multirow{28}{*}{No} & GeForce  GTX 580 & FP32 & 1.58 & 244 & 09/11/2010 & Desktop & 6.48 \\ 
 & GeForce GTX 590 & FP32 & 2.49 & 365 & 24/03/2011 & Desktop & 6.82 \\
 & GeForce GTX 680 & FP32 & 3.09 & 195 & 22/03/2012 & Desktop & 15.85 \\
 & GeForce GTX 690 & FP32 & 5.62 & 300 & 29/04/2012 & Desktop & 18.73 \\
 & Tesla K10 & FP32 & 4.58 & 225 & 01/05/2012 & Server & 20.36 \\
 & Tesla K20x & FP32 & 3.94 & 235 & 12/11/2012 & Server & 16.77 \\
 & GeForce GTX 780 & FP32 & 4.16 & 250 & 23/04/2013 & Desktop & 16.64 \\
 & Tesla K40 & FP32 & 5.04 & 235 & 08/10/2013 & Server & 21.45 \\
 & GeForce GTX 780 TI & FP32 & 5.35 & 250 & 07/11/2013 & Desktop & 21.40 \\
 & GeForce GTX Titan Black & FP32 & 5.65 & 250 & 18/02/2014 & Desktop & 22.60 \\
 & GeForce GTX Titan Z & FP32 & 8.12 & 375 & 28/05/2014 & Desktop & 21.65 \\
 & GeForce GTX 980 & FP32 & 4.98 & 165 & 18/09/2014 & Desktop & 30.18 \\
 & Tesla K80 & FP32 & 8.22 & 300 & 17/10/2014 & Server & 27.40 \\
 & GeForce GTX TITAN X & FP32 & 6.69 & 250 & 17/03/2015 & Desktop & 26.76 \\
 & GeForce GTX 980 Ti & FP32 & 6.06 & 250 & 02/06/2015 & Desktop & 24.24 \\
 & Tesla M60 & FP32 & 9.65 & 300 & 30/08/2015 & Server & 32.17 \\
 & Tesla M40 & FP32 & 6.84 & 250 & 10/10/2015 & Server & 27.36 \\
 & GeForce GTX 1080 & FP32 & 8.87 & 180 & 26/05/2016 & Desktop & 49.28 \\
 & TITAN X Pascal & FP32 & 10.97 & 250 & 02/08/2016 & Desktop & 43.88 \\
 & GeForce GTX 1080 Ti & FP32 & 11.34 & 250 & 10/03/2017 & Desktop & 45.36 \\
 & TITAN XP & FP32 & 12.15 & 250 & 06/04/2017 & Desktop & 48.60 \\
 & Tesla V100 & FP32 & 15.70 & 300 & 27/03/2018 & Server & 52.33 \\
 & Tesla T4 & FP32 & 8.10 & 70 & 13/09/2018 & Server & 115.71 \\
 & GeForce RTX 2080 & FP32 & 10.07 & 215 & 20/09/2018 & Desktop & 46.84 \\
 & GeForce RTX 2080 Ti & FP32 & 13.45 & 250 & 20/09/2018 & Desktop & 53.80 \\
 & Nvidia Titan RTX & FP32 & 16.31 & 280 & 18/12/2018 & Desktop & 58.25 \\
 & GeForce RTX 3080 & FP32 & 29.80 & 320 & 01/09/2020 & Desktop & 93.13 \\
 & GeForce RTX 3090 & FP32 & 35.60 & 350 & 01/09/2020 & Desktop & 101.71 \\ \midrule
 \multirow{4}{*}{For CNN} & Tesla V100 & Mixed & 35.71 & 300 & 27/03/2018 & Server & 119.03 \\
 & Tesla T4 & Mixed & 21.85 & 70 & 13/09/2018 & Server & 312.15 \\
 & A100 & TF32 & 27.41 & 400 & 14/04/2020 & Server & 68.52 \\
 & A100 & Mixed & 52.35 & 400 & 14/04/2020 & Server & 130.88 \\ \midrule
 \multirow{4}{*}{For NLP} & Tesla V100 & Mixed & 41.44 & 300 & 27/03/2018 & Server & 138.13 \\
 & Tesla T4 & Mixed & 25.58 & 70 & 13/09/2018 & Server & 365.46 \\
 & A100 & TF32 & 55.85 & 400 & 14/04/2020 & Server & 139.64 \\
 & A100 & Mixed & 73.29 & 400 & 14/04/2020 & Server & 183.23 \\ \bottomrule

\end{tabular}%
}
\caption{GPUs throughput and power consumption data compilation.}
\label{tab:gpu_models}
\end{table}

\end{document}